\ifdefined\pdfoutput\ifdefined\XeTeXrevision\else\pdfoutput=1\fi\fi
\PassOptionsToPackage{unicode}{hyperref}
\PassOptionsToPackage{hyphens}{url}
\PassOptionsToPackage{dvipsnames,svgnames,x11names}{xcolor}
\documentclass[
  11pt,
]{article}
\usepackage{xcolor}
\usepackage[margin=1in]{geometry}
\usepackage{amsmath,amssymb}
\usepackage{iftex}
\ifPDFTeX
  \usepackage[T1]{fontenc}
  \usepackage[utf8]{inputenc}
  \usepackage{textcomp} 
\else 
  \usepackage{unicode-math} 
  \defaultfontfeatures{Scale=MatchLowercase}
  \defaultfontfeatures[\rmfamily]{Ligatures=TeX,Scale=1}
\fi
\usepackage{lmodern}
\ifPDFTeX\else
\fi
\IfFileExists{upquote.sty}{\usepackage{upquote}}{}
\IfFileExists{microtype.sty}{
  \usepackage[]{microtype}
  \UseMicrotypeSet[protrusion]{basicmath} 
}{}
\makeatletter
\@ifundefined{KOMAClassName}{
  \IfFileExists{parskip.sty}{%
    \usepackage{parskip}
  }{
    \setlength{\parindent}{0pt}
    \setlength{\parskip}{6pt plus 2pt minus 1pt}}
}{
  \KOMAoptions{parskip=half}}
\makeatother
\usepackage{longtable,booktabs,array}
\usepackage{calc} 
\usepackage{etoolbox}
\makeatletter
\patchcmd\longtable{\par}{\if@noskipsec\mbox{}\fi\par}{}{}
\makeatother
\IfFileExists{footnotehyper.sty}{\usepackage{footnotehyper}}{\usepackage{footnote}}
\makesavenoteenv{longtable}
\usepackage{graphicx}
\makeatletter
\newsavebox\pandoc@box
\newcommand*\pandocbounded[1]{
  \sbox\pandoc@box{#1}%
  \Gscale@div\@tempa{\textheight}{\dimexpr\ht\pandoc@box+\dp\pandoc@box\relax}%
  \Gscale@div\@tempb{\linewidth}{\wd\pandoc@box}%
  \ifdim\@tempb\p@<\@tempa\p@\let\@tempa\@tempb\fi
  \ifdim\@tempa\p@<\p@\scalebox{\@tempa}{\usebox\pandoc@box}%
  \else\usebox{\pandoc@box}%
  \fi%
}
\def\fps@figure{htbp}
\makeatother
\providecommand{\tightlist}{%
  \setlength{\itemsep}{0pt}\setlength{\parskip}{0pt}}

\usepackage{amsmath,amssymb}
\usepackage{booktabs}
\usepackage{float}
\usepackage{url}
\makeatletter
\g@addto@macro\UrlBreaks{\do\.\do\-\do\_\do\/\do\:\do\=\do\?\do\&}
\makeatother
\usepackage{iftex}
\ifPDFTeX
  \usepackage[LGR,T1]{fontenc}
  \usepackage{textalpha}
\fi
\newcommand{\gk}[1]{#1}
\usepackage{bookmark}
\IfFileExists{xurl.sty}{\usepackage{xurl}}{} 
\hypersetup{
  pdftitle={Metaphor Tracer: A Theory-Informed Analysis of Hidden States},
  pdfauthor={Marc Heimann, Roxana Assadi Moghaddam, Olga Brovkina, Mark Pettifor, Lutz Götzmann},
  colorlinks=true,
  linkcolor={blue},
  filecolor={Maroon},
  citecolor={Blue},
  urlcolor={blue},
  pdfcreator={LaTeX via pandoc}}

\title{Metaphor Tracer: A Theory-Informed Analysis of Hidden States}
\author{\parbox[t]{\dimexpr\textwidth-2\tabcolsep\relax}{\centering Marc~Heimann, Roxana~Assadi~Moghaddam, Olga~Brovkina, Mark~Pettifor, Lutz~Götzmann}}
\date{}

\begin{document}
\maketitle

\ensuremath{^{1}} Hermeneutic AI. * Corresponding author:
marc@hermeneutic.ai

\textbf{Author contributions.} Marc Heimann conceived and designed the
study, developed its conceptual framework, implemented the full pipeline
and analysis, and wrote the manuscript; he is responsible for the work
as a whole and is the corresponding author. Roxana Assadi Moghaddam
contributed to the conceptual framing and, as the annotating analyst,
produced the psychoanalytic annotations of the clinical transcripts.
Olga Brovkina served as the annotating scientist for the domain-table
material, designed the table test, and provided app testing and
feedback. Mark Pettifor designed the engineered texts and provided app
testing and feedback. Lutz Götzmann contributed to the conceptual
development and provided access to the patient data.

\textbf{Abstract:} What do a language model's hidden states say about
the organization of a single text? From one forward pass, without
training, we score every token position on two properties. The
\emph{aggregator} measures whether the position consolidates the whole
text into a stable configuration. The \emph{differentiator}, whether
other tokens are transiently carried into its subspace as the model
reads: metaphor in its root sense, transport. Constants were frozen on
one discovery text; every other is confirmatory.

The aggregator is not, in the classic sense, an information measure, nor
a measure of salience. Across three unrelated models, as a signifier
repeats, its surprisal and its attention drain while its aggregator
score holds: the channel marks a token's place in the text. That this
tracks a reading rests on independent ground truth: an engineered
register the aggregator follows across its boundaries (6/6 cells), and a
psychoanalyst's marking of clinical transcripts, fixed before the
instrument existed, in 34/36 cells, with a graded increment above
lexical controls and dissociations no type-level measure reproduces. A
transfer test gives the result its shape: the model whose token
structure travels with lexical type reads the singular discourse worst,
and in a matched base/instruct pair tuning raises fidelity without
moving type-transfer. Structural value is a property of a token's place
in \emph{this} text, not of its vector alone: a relational rather than
essentialist reading of hidden states, operationalizing theory that
predated the instrument.

\section{Introduction}\label{introduction}

One of the most basic concepts of the way the humanities study text is
the understanding that a text's individual meaning is inherent to it and
only it. It is partially derived and certainly influenced by a preceding
history, and it may even be understood by the consequences it produced,
but fundamentally it is impossible to speak about, for example,
\emph{Les Chants de Maldoror} without approaching the inherent logic of
this specific text. Consequently, every text is an individual one.
Similarly, a large language model, reading a text, organizes it as such
an individual geometry. Some positions come to anchor the whole, for
example, a heading, a recurring name, a closing period that gathers a
paragraph behind it, while others are merely passed through. This
organization is not the model's output and also not a decision to be
explained as is common in explainable AI approaches; it is the specific
and individual shape of a particular reading, laid down in the residual
stream during a single forward pass. This individual reading, its
geometry of internal references, can be made visible. We build a
training-free instrument that reads that shape back out, and allows
testing what it reads against what human experts independently mark.

The instrument did not come first. In prior theoretical work one of the
authors argued, on the shared principles of Lacanian theory and
transformer architecture, that the ``understanding'' a language model
performs must have a describable structure: points that quilt a
discourse and fix its meaning retroactively (the \emph{point de
capiton}, the master signifier), an engine of transport by which a term
is carried into a region not its own (metaphor), and the patterned
structure this engine operates on (metonymy, not operationalized here,
as the basic idea of a signifier determined only by its position in
relation to other signifiers is arguably already realized in vector
embeddings, which possess no referent) (Heimann and Hübener 2024,
further developed in Heimann and Hübener 2025 and Heimann 2026a--c).
That argument had no measuring device; it was built on a classical
philosophical method: the comparison of principles, architectural and
mathematical. The present paper is a first operationalization of these
insights: we construct the device, freeze it on a single discovery text
before any confirmatory run, and ask whether the structures the theory
named are present, replicable, and aligned with expert judgment. The
theoretical claims of alignment are thus made falsifiable here, and, as
we report, they survive contact with the geometry while acquiring
precise mechanical content. What the pipeline tests is accordingly not a
practical use but, first and foremost, an interpretation: the model read
as a theory of language in the tradition of Freudian thought, with the
geometry as the place where this reading becomes legible in form. This
makes no claim about the correctness of the theory as such, but shows
that it can inform varied forms of explainable-AI methods by binding
them together into a coherent interpretation. Thus no new analysis
primitive is introduced here; what is new is the composition, existing
tools thought together as a method in which several inform each finding,
and the object they are pointed at. In view of the existing, often
implicit underpinnings of explainable-AI studies, this also presents a
counter-reading, as this approach owes no fealty or loyalty to the
analytic philosophy discourse dominant in computer science.

The instrument scores every token position on two mechanistically
distinct properties: an \emph{aggregator} that consolidates the text
into a configuration around an anchor, and a \emph{differentiator} that
registers transport of other tokens into that anchor's subspace, each
read in two views of an anisotropic space (Methods). It reads at
resolution rather than in aggregate: 13 carefully curated texts yield a
diagnostic database of over 40 gigabytes, the raw float32 activation
tensors and the per-anchor subspace geometry computed from them, so that
each reading event is scanned along its whole spatial-depth trajectory
instead of being averaged away over a corpus of low-resolution
statistics.

The paper proceeds as follows. The theoretical framework places the
reading beside hidden-state interpretability, and Methods define the
anchor, the two channels, the two views and the frozen battery. Results
run from engineered ground truth, where the construction is known
(§1--§4), to two expert ground truths of different kinds: a
psychoanalyst's marking of clinical transcripts (§5) and a domain
expert's column declaration on execution traces (§7). Between them sits
the transfer test that gives the result its shape (§6), and §8 collects
the robustness checks. The Discussion takes up what the channels
measure, what an expert annotation can and cannot certify, and what a
single-annotator corpus of this size leaves open.

\textbf{Contributions.} (I) A complementary perspective in which
existing hidden-state methods can be seen as analyzing one and the same
object: the large language model as a theory of language in the
tradition of Freudian thought. This makes prior theoretical claims
testable against model internals. (II) A training-free, single-pass
instrument that reads the relational structure of an individual text
from hidden-state geometry, with two mechanistically distinct channels
and two anisotropy views. (III) A frozen confirmatory battery over
engineered and expert ground truth, with results reported as replication
counts across model\ensuremath{\times}text\ensuremath{\times}view cells rather
than per-cell significance. (IV) A first probe, to our knowledge, of
model-internal structure against psychoanalytic expert annotation: one
ground truth among several, single-annotator, and reported with the
lexical controls it requires. (V) A relationality result: structural
value is measurably less stable than the state it is read from, and, in
the models that read the singular discourse best, does not transfer with
lexical type, operationalizing a Saussurean/Lacanian anti-essentialism
as a testable property of transformer representations. Throughout, we
hold the work to rigor in both registers, continental-theoretic and
mechanical, and mark for every finding which view and which channel
carries it.

\section{Theoretical Framework}\label{theoretical-framework}

A survey of neighboring literatures has, in this paper, an unusual
second task, although one common to philosophical literature: it must
explain a part of the methodology that method sections do not normally
discuss. The main author of this paper is a philosopher, not a computer
scientist, and what the paper offers is accordingly not primarily a
measurement but a metric informed by theory, one that is found precisely
because we are looking for the metaphoric in Lacan's sense (below), and
not for a property of words. The complementary perspective set out above
follows from it: existing methods seen as analyzing one and the same
object, the model read in the Freudian tradition. Despite published
groundwork (Heimann and Hübener 2024, 2025; Heimann 2026a, 2026b,
2026c), the combination remains a highly unusual one, and it is what
made the instrument buildable, letting the first author devise a
pipeline that tests an interpretation rather than a practical use.

Concretely, the pipeline sits at the intersection of several
literatures, none of which, to our knowledge, combines more than two of
its defining features: training-free single-pass hidden-state geometry;
per-text relational anchor structure rather than corpus aggregates; dual
anisotropy views treated as complementary maps; two channels separating
processing from final-state properties; a frozen confirmatory battery
with engineered and expert ground-truth tiers; and validation against
psychoanalytic expert annotation. Below we name the nearest neighbors
and the distinctions that we introduce. The purpose of the survey,
however, is integrative: read from the perspective above, these
literatures (geometric, mechanistic, evaluative, clinical) are partial
analyses of a common object, and what the paper contributes is the frame
under which they become commensurable.

The closest methodological cousin studies the geometry of tokens in LLM
internal representations, intrinsic dimension, neighborhood overlap,
per-layer cosine over a large corpus, correlating geometry with
token-prediction loss (Viswanathan et al.~2025). The analysed raw
material is the same (per-token residual-stream geometry across depth);
the question, however, is orthogonal, being corpus-level statistics
predicting loss, with no anchor construction and no ground truth about
the structure of an individual text. Layer-wise participation-ratio and
effective-rank profiles likewise appear as global representation-quality
diagnostics, not as per-anchor subspace measures. Also, a substantial
literature documents attention sinks (Peng et al.~2026). SepLLM (Chen et
al.~2024), for example, compresses a segment into a single separator
token, establishing that a separator can carry its segment's content
well enough to substitute for it, which is the precondition for a
delimiter to hold a pin function at all. It is not, however, the
explanation of our pins: within the delimiter class the positions our
aggregator ranks highest are, in the instruction-tuned models, the
attention-\emph{poor} ones (Results §8), so whatever makes delimiters
preferential aggregation sites is not what selects which delimiter binds
which segment. This work explains why delimiter anchors are available
and how they can be used in practice; none of it reads them back as a
map of a text's structure, which is our object. In regard to the
anisotropy mapping, Timkey and van Schijndel (2021) show that removing
rogue high-variance dimensions aligns model similarity spaces better
with human similarity judgments; related work finds outlier dimensions
to be task- or knowledge-relevant (Rudman and Eickhoff 2023; Rudman et
al.~2023). Consequently, the field's standard move is to correct
anisotropy. Our two-view design instead retains both geometries and
treats their disagreement as signal, the boundary alignment that inverts
between views, the complementary structure each carries. We find no
precedent for this dual-view-as-dual-map use.

Central to the approach is the search for metaphoricity as Lacanian
theory conceives it: metaphor as an operation that rips the signifier
from its lexical linkages and reassociates it via identification (Lacan
1993, 218--19). We do find a substantial body of research on metaphors
in LLMs. Supervised probing locates metaphoricity in the middle layers
of pre-trained models against VUA-style word annotations, with
cross-lingual transfer (Aghazadeh et al.~2022); more recent work
identifies metaphor via prompting or fine-tuning (Fuoli et al.~2025).
There is a notable discourse on metaphor as an instrument of
jailbreaking and prompt efficiency (Kramer 2025; Yan et al.~2025).
However, the linguistic understanding of this discourse is mostly that
of a classification against word-level labels. Ours is unsupervised and
geometric, and while it looks for operations of metaphoricity, it
separates processing metaphoricity (transport in flight, the red
channel) from final-state metaphoricity, a distinction the
classification framing does not draw and one our correlational setup
deliberately does not push beyond the processing side.

To test this not only against engineered texts but also against expert
language users, our evaluation protocol, the rank AUC of channel values
on expert-marked versus unmarked tokens, follows the plausibility
tradition in XAI (Kazmierczak et al.~2024; Jahromi et al. 2024), which
we cite as its home. However, the object differs categorically: there is
no task and no classifier decision. The pipeline explains a reading, not
a decision, and the ground truth is where an expert locates organizing
structure, not which tokens justify a label. Downstream, the boundary
test is a segmentation capability, adjacent to the TextTiling lineage
and its sentence-embedding successors (Jia and Diaz-Rodriguez 2026;
Solbiati et al.~2021). But segmentation is a consequence, not the
object: none of these methods expose channels, views, anchors, or a
processing/consolidation distinction, and none target the within-text
organization the channels read. Lastly, work estimating clinical
constructs from transcripts is uniformly output-based, prompted or
fine-tuned scoring of content (Abdou et al.~2025). There is also a large
corpus of literature applying psychological and psychoanalytic methods
to large language models, which we ignore (e.g.~Magee et al.~2023). The
reason is simple: these are output-focused or remain at the level of
theory, the latter including work that takes up our own earlier
framework (Geal 2025, building on Heimann and Hübener 2024). None to our
knowledge reads hidden-state geometry, and none is validated against
psychoanalytic annotation of signifiers and ruptures.

Concurrent with our work, Gurnee et al.~(2026) identify, with a causal
corpus-averaged instrument (the Jacobian lens), a sparse set of
verbalizable directions that behave like a global workspace. The
instruments share nothing methodologically, and we develop the
comparison, the structural convergences, the depth divergence, and the
question of which vocabulary imports less, in the Discussion.

We claim priority for the perspective and for its operationalization:
for the combination of the six features above and, specifically, for
validation against psychoanalytic expert annotation, against the
neighbors named here. We deliberately avoid any claim that no comparable
explanation methods exist: the plausibility literature above is exactly
comparable in protocol and differs in object rather than in kind. What
we could find no neighbor for is the object itself: the model treated
not as a tool whose outputs want explaining, but as a theory of language
to be read.

\section{Methods}\label{methods}

\subsection{Models and extraction}\label{models-and-extraction}

We analyze three causal transformer language models:
\texttt{microsoft/\allowbreak{}phi-\allowbreak{}4} (14B; 40 layers, d = 5120),
\texttt{Qwen/\allowbreak{}Qwen3-\allowbreak{}8B} (8B; 36 layers, d = 4096), both
instruction-tuned, and \texttt{meta-\allowbreak{}llama/\allowbreak{}Llama-\allowbreak{}3.\allowbreak{}1-\allowbreak{}8B} (8B; 32 layers,
d = 4096), a base model with no instruction tuning. The set is chosen
for diversity of architecture, tokenizer, and training regime; with
three models we make no factorial claims about size or alignment, though
we note where the base model diverges from the two tuned models.

A fourth run set enters as a \emph{matched-lineage follow-up}:
\texttt{meta-\allowbreak{}llama/\allowbreak{}Llama-\allowbreak{}3.\allowbreak{}1-\allowbreak{}8B-\allowbreak{}Instruct}, the instruction-tuned twin of
the base model, over every corpus text, same pipeline, same frozen
constants, same annotations. Its results are reported separately
wherever they appear (Results §1, §3, §5 and §7) and are gathered in §6;
they enter no pooled statistic or replication count. Its single purpose
is to separate what instruction tuning changes from what the lineage
fixes, which the three-model corpus cannot do on its own. The checkpoint
is public and the battery released, so these follow-up numbers
regenerate by the same procedure as the rest of the corpus, the same two
model passes (state extraction; horizon logits) over the named
checkpoint.

All analyses consume a single teacher-forced forward pass per text in
float32, with no sampling; given model and text the procedure is
deterministic. For a text of T tokens we retain the full hidden-state
tensor \(h \in \mathbb{R}^{L \times T \times d}\), where \(h_\ell(t)\)
is the residual-stream state of token t after transformer block
\ensuremath{\ell}. The embedding layer is excluded: \ensuremath{\ell} = 1
denotes the first block's output and \ensuremath{\ell} = L the final
layer. Each extraction records its provenance (model id and tensor
shape, \texttt{run\_\allowbreak{}meta.\allowbreak{}json}); run folders are keyed by text
\emph{and} model, and cached scan artifacts are refused when states have
been re-extracted over them.

\textbf{Token pool.} All anchor statistics draw on a token pool that
excludes special tokens (\texttt{\textless{}\textbar{}.\allowbreak{}.\allowbreak{}.\allowbreak{}}) and bare
tokenizer space-markers, but deliberately retains punctuation: delimiter
tokens are documented aggregation sites in transformers, and excluding
them would remove precisely the positions where segment-level binding is
hypothesized to occur. (This choice is consequential: in phi-4 the top
aggregator anchors are almost exclusively paragraph-final delimiters;
see Results §3, and §8, where the pins dissociate from attention.)
Excluded tokens enter no anchor statistic: both the member and
non-member means of the differentiator (below) are taken over pool
tokens only.

\subsection{Anchor construction}\label{anchor-construction}

Every scanned token position a is evaluated as a candidate
\emph{anchor}. For each dimension j we compute the median, over all pool
tokens i \ensuremath{\neq} a, of the growth of the elementwise product
with the anchor between the first and final layer:

\[\Delta_j(a) = \operatorname*{median}_{i}\bigl[\, h_L(i)_j \cdot h_L(a)_j - h_1(i)_j \cdot h_1(a)_j \,\bigr].\]

Sorting \ensuremath{\Delta}(a) in descending order, the anchor's \emph{core
subspace} dims(a) is the smallest prefix of dimensions whose cumulative
sum covers a fraction \ensuremath{\mu} = 0.7 (\texttt{mass\_\allowbreak{}cut}) of the
\emph{positive} mass \(\sum_j \max(\Delta_j, 0)\). The restriction to
positive mass is essential: the signed distribution is dominated by
cancellation, and normalizing by the net total collapses the subspace
degenerately (documented in the implementation, \texttt{mass\_\allowbreak{}cut\_\allowbreak{}k}).
The subspace size k = \textbar dims(a)\textbar{} varies over orders of
magnitude between anchors and between models (median k per text: single
digits in Qwen3-8B, single digits to hundreds in phi-4, hundreds in the
base model; heavy upper tail throughout) and is itself an informative
statistic.

\subsection{Depth recruitment and
basins}\label{depth-recruitment-and-basins}

At each layer \ensuremath{\ell}, token i is \emph{recruited} by anchor a
when the cosine similarity between \(h_\ell(i)\) and \(h_\ell(a)\),
restricted to dims(a), reaches a threshold \(\tau_\ell\). The
\emph{recruitment layer} of i (\texttt{recruited\_\allowbreak{}layer}) is the first
such \ensuremath{\ell}; if the similarity later falls below threshold, the
layer of loss is recorded (\texttt{lost\_\allowbreak{}layer}).

The anchor's \emph{basin} is the \emph{ever-recruited} set: every token
that crossed the threshold at any depth, whether or not it remains above
it at the final layer. This is a deliberate definition, not a
convenience. Recruitment is typically a mid-depth event whose
final-layer trace is partial. For strong aggregator anchors, basin
membership saturates well before the final layer (mid-stack in some
models, late-stack in others), while final-layer cosines settle far
below threshold. The basin therefore describes what the model's reading
\emph{passes through}, not the state it comes to rest in; the channels
built on it (next subsection) inherit this processing character.

\begin{figure}[H]
\centering
\pandocbounded{\includegraphics[keepaspectratio,alt={Figure 1 --- anchor construction}]{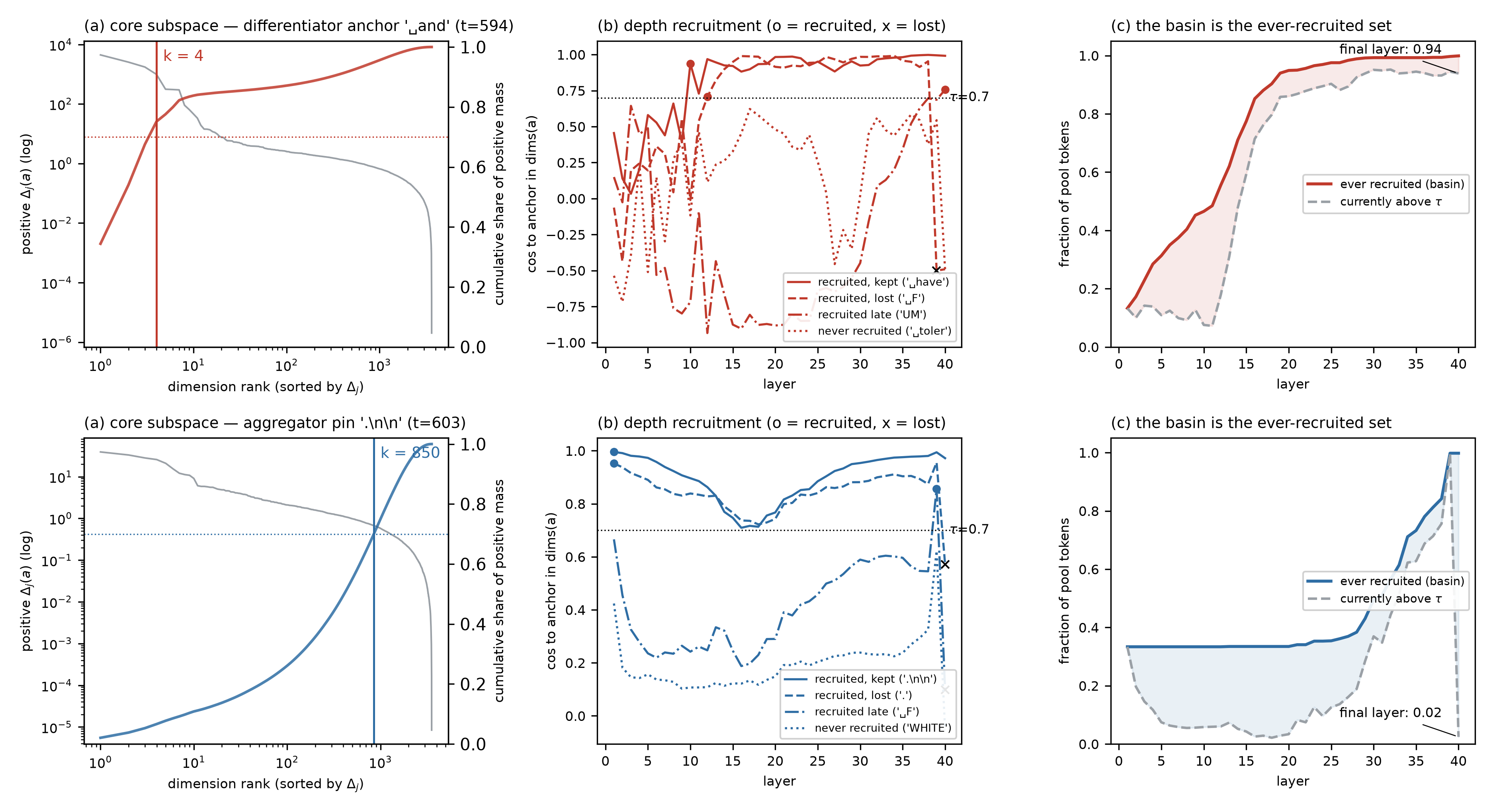}}
\end{figure}

\textbf{Figure 1 --- anchor construction, on cached artifacts of one
run} (a top differentiator anchor and a top aggregator pin of the
discovery text, phi-4). (a) The sorted positive interaction growth
\ensuremath{\Delta} with the \ensuremath{\mu} = 0.7 cut fixing the core
subspace; (b) per-token cosine trajectories with the \ensuremath{\tau} =
0.7 threshold, recruitment and loss marked; (c) the ever-recruited basin
against currently-above-threshold membership per layer --- the gap that
gives the channels their processing character.

\subsection{Two channels}\label{two-channels}

Each anchor is scored on two mechanistically distinct properties, and
both are properties of metaphoricity in the root sense of the word:
metaphor, from \gk{μετά} (\emph{metá}, ``across, beyond'') +
\gk{φέρειν} (\emph{phérein}, ``to carry, bear''), transport, a
term carried into a region not its own. One channel marks the transport
in flight, the other what it consolidates into.

\textbf{Differentiator strength --- a mark of processing metaphoricity}
(rendered red). For every token position i let \(p_i(a)\) be the
fraction of its depth-displacement energy falling inside the anchor's
core subspace (\texttt{pull\_\allowbreak{}frac}):

\[p_i(a) = \sum_{j \in \mathrm{dims}(a)} \bigl(h_L(i)_j - h_1(i)_j\bigr)^2 \big/ \bigl\lVert h_L(i) - h_1(i) \bigr\rVert^2.\]

The \emph{pull gap} (\texttt{pull\_\allowbreak{}frac\_\allowbreak{}gap}) is the mean of \(p_i(a)\)
over basin members minus its mean over non-members; differentiator
strength is max(0, pull gap) \ensuremath{\cdot} \(p_a(a)\)
(\texttt{self\_\allowbreak{}pull\_\allowbreak{}frac}). Where the basin is empty or exhausts the
pool the gap is undefined and the strength is set to zero, a tie mass
acknowledged by the midrank percentiles used wherever red percentiles
are reported. It is high when the tokens the anchor ever recruited moved
distinctively into the anchor's subspace, an anchor that divides the
text into an inside and an outside during processing. Because the basin
is the ever-recruited set, this channel does \emph{not} measure a
property of the final representation: it marks positions at which
depth-transient alignment (metaphoric transport in flight) occurred
somewhere in the model's reading. We accordingly interpret red as
marking \emph{processing} metaphoricity: \texttt{pull\_\allowbreak{}frac} measures
how much of a token's journey through the stack was spent moving into
someone else's territory. We defer claims about final-state
metaphoricity to interventions outside this paper's scope. The quantity
is also \emph{relational}: it depends on the whole token pool, not on
the anchor alone.

\textbf{Aggregator strength} --- a mark of consolidating metaphoricity
(rendered blue). The channel is the participation ratio
(\ensuremath{\Sigma}\ensuremath{\sigma})\ensuremath{^{2}}/\ensuremath{\Sigma}\ensuremath{\sigma}\ensuremath{^{2}}
of the singular values of the final-layer configuration of pool tokens,
centered over tokens and restricted to dims(a), normalized by the hidden
dimension d (\texttt{effective\_\allowbreak{}rank}, \texttt{capacity\_\allowbreak{}share}). It is
high when the anchor's subspace supports a broad, high-rank
configuration of the whole text; it is read as a consolidation point
rather than a divider. Unlike the red channel, this \emph{is} a
final-layer property. Transparency note: across tested texts,
capacity\_share is strongly rank-correlated with the subspace size k
(Spearman 0.85--1.00 across runs, median \ensuremath{\approx} 0.98,
\ensuremath{\approx} 0.99 in phi-4); why the channel measures occupancy
rather than a bare dimension count, despite this coupling, is treated in
the Limitations.

For visualization, each channel is contrast-normalized within the
scanned tokens of one run (median \ensuremath{\mapsto} 0, 95th percentile
\ensuremath{\mapsto} 1, clipped); raw values are retained in all outputs.
\emph{Strong} anchors are the union of the top decile of each channel
(\texttt{STRONG\_\allowbreak{}QUANTILE\ =\ 0.\allowbreak{}90}). Normalization scope never exceeds
the scope of the question: within-text for single-text maps, pooled
across variants only where the question is itself comparative (prompt
comparison), and never across models.

\subsection{Two views: operative and
substrate-partialled}\label{two-views-operative-and-substrate-partialled}

Transformer residual streams are strongly anisotropic: a small set of
high-variance dimensions dominates inner products and is shared across
the whole text. These dimensions are not noise (the model's attention
arithmetic runs on them), so we analyze two views and treat their
\emph{disagreement} as informative:

\begin{enumerate}
\def\labelenumi{\arabic{enumi}.}
\tightlist
\item
  \textbf{Operative view.} Raw states; fixed \ensuremath{\tau} = 0.7 at
  every layer. The geometry in which the model actually computes.
\item
  \textbf{Quiet-dimensions view.} Every dimension z-scored per layer
  over the text's tokens before scanning; \(\tau_\ell\) set per layer at
  the 95th percentile of that layer's cosine distribution (fixed
  thresholds are calibrated to the anisotropic raw space and do not
  transfer). This partials out the dominant-variance substrate and
  exposes structure in low-variance dimensions.
\end{enumerate}

The views dissociate strongly and consistently: on engineered texts the
aggregator channel's boundary alignment \emph{inverts} between views,
and per-anchor values decorrelate (Results). We report for every finding
which view carries it.

\subsection{Directional split of anchor
growth}\label{directional-split-of-anchor-growth}

Because the model is causally masked, tokens before an anchor cannot
have been influenced by it. For each anchor we therefore split
\ensuremath{\Delta}(a), restricted to dims(a), into per-dim medians pooled
over tokens before vs.~after the anchor position, reporting each side's
positive mass and their ratio (\texttt{delta\_\allowbreak{}before\_\allowbreak{}mass},
\texttt{delta\_\allowbreak{}after\_\allowbreak{}mass}, \texttt{before\_\allowbreak{}share}; sides with fewer
than 8 pool tokens yield NaN). Before-mass reads unambiguously as the
anchor aligning to (absorbing) its past; after-mass is a candidate for
forward transmission and remains correlational. We note a measured
limitation: the per-dim median is a statistic of the \emph{typical}
token, which is substrate-dominated and side-invariant, so this split is
a coarse instrument; the token-resolved binding test of the battery
(paragraph binding, below) is the sensitive one.

\subsection{Profiles}\label{profiles}

\textbf{Consolidation profile.} The rolling density (centered window w =
max(20, T/15)) of strong anchors across token position. Low density:
token movement is absorbed by established structure; high density: the
text is still erecting organizing points.

\textbf{Horizon profile.} The Shannon entropy of the full next-token
distribution at each position. This requires logits, which the state
extraction (a bare encoder pass) does not produce; entropies come from a
\emph{second deterministic teacher-forced pass} through the same model
with its LM head, over the identical token sequence reconstructed from
the stored tokens; positions therefore align with the scan exactly, and
a vocabulary check refuses token sequences from a different model. The
same pass yields the per-position \emph{surprisal} of the realized
token, \(-\log p(x_t \mid x_{<t})\) (NaN at the unconditioned first
position).

\textbf{Novelty baseline.} A critical reading could hold that anchor
channels merely track how difficult the text is. Wherever a horizon
profile exists, the battery therefore reports the token-level Spearman
correlation of each channel's raw values against horizon entropy and
against realized surprisal, per view; the outcome (the channels are not
reducible to either) is reported in Results §8.

\textbf{Mundane baselines.} A second critical reading could hold that
the analyst's marks are recoverable from a word list. On the clinical
tier the battery therefore scores static lexical predictors on the same
marked/unmarked split and the same rank statistic: inverse corpus
frequency (Zipf, \texttt{wordfreq}), word length, content-word
membership, term frequency and TF-IDF, word-level surprisal, predictive
entropy, and an unfitted rank-combination of content, rarity and
surprisal. None has a fitted parameter; word-level features project onto
tokens by the same span-overlap rule the marks use, and non-word tokens
receive each feature's least mark-predictive value, so the projection
cannot manufacture separation. The aggregator is then rescored within
content words, where the content/function composition is constant by
construction, and rank-residualized on word-level surprisal and rarity.
Its coupling to those covariates is measured separately, over all
scanned tokens and without the annotation entering, as the share of its
rank variance they explain. Two within-transcript nulls calibrate the
remainder, 10,000 seeded draws each: a block null re-placing a
transcript's marked token blocks at random, count and lengths preserved,
and a rarity-matched null drawing token sets to match the marks'
within-run rarity-decile histogram. The same predictors are scored on
the two region grounds by the ranking their channel results use
(\texttt{baseline\_\allowbreak{}aucs.\allowbreak{}py}, \texttt{baseline\_\allowbreak{}regions.\allowbreak{}py}; Results §4,
§5).

\textbf{Attention-saliency pass.} A post-hoc control on the clinical
tier (Results §8) scores the standard model-internal saliency reading on
the same marked/unmarked split: per-token \emph{attention received}
(head- and layer-averaged column mass) and \emph{attention rollout}
(Abnar and Zuidema 2020; per layer the head-mean matrix mixed with the
residual path, 0.5\ensuremath{\cdot}Ā + 0.5\ensuremath{\cdot}I, composed
bottom-up), together with the final position's rollout row. Weights come
from a third teacher-forced pass over the identical reconstructed token
sequence, with the same position alignment and vocabulary check as the
horizon pass. Two rules were fixed before any value was computed: every
received-mass measure is divided by the token's number of attending
queries, since under a causal mask early positions accumulate mass
mechanically; and each measure is also scored within content words and
residualized on the lexical covariates, the same restrictions the
aggregator channel receives. This pass alone runs in bfloat16 with eager
attention (the full attention tensor does not fit beside float32
weights); it feeds only rank statistics, and no frozen battery number
depends on it (\texttt{attention\_\allowbreak{}saliency.\allowbreak{}py}; saliency profiles in a
parallel output tree, the frozen run directories untouched).

\subsection{Minimal-pair (A/B) protocol}\label{minimal-pair-ab-protocol}

For text pairs differing by one contiguous inserted passage (declared in
the manifest with the passage's first and last sentence as markers), the
edit's controlled shape makes alignment exact: shared prefix and shared
suffix correspond character-for-character, so no sequence alignment is
needed; tokens that BPE merges differently at the junctions simply fail
the correspondence and are dropped. Three measurements follow
(\texttt{minimal\_\allowbreak{}pair.\allowbreak{}py}; state-level results are cached per pair and
invalidated when either run is re-extracted):

\textbf{Prefix identity floor.} Prefix tokens are verified
string-identical, and the final-layer cosine between the two runs'
prefix states is reported (minimum over tokens). Causal masking predicts
exact identity; the observed floor (1.0 in float32 across all models)
calibrates every other similarity in the protocol.

\textbf{Suffix divergence.} The shared suffix is identical wording read
after different pasts. We report per-token final-layer cosine between
versions as a function of distance from the edit (median over the first
and last 20 aligned tokens), and alongside it \emph{channel stability}:
the Spearman rank correlation of each channel's raw values over aligned
suffix tokens. The channels are relational text-level measures (basins,
subspaces, and quantiles are computed against the whole token pool), so
their stability can fall below that of the states they are read from;
the outcome (Results §4) quantifies this designed context-sensitivity,
and findings are accordingly claimed at the level of within-text
structure, never as token-local absolute values.

\textbf{Junction specificity.} The edit region's strong-anchor density
ratios in the edited text are compared with a token-count-matched window
at the base text's junction. A signature that appears in the edit region
but not at the matched junction belongs to the \emph{content} of the
passage, not to its position in the document.

\subsection{Corpus design and the frozen test
battery}\label{corpus-design-and-the-frozen-test-battery}

The corpus has two tiers of ground truth. \textbf{Tier 1 --- engineered
texts}: the boundary between structural parts is known to the character
from the construction recipe; boundary markers and pre-registered
directional predictions are fixed in a manifest
(\texttt{battery\_\allowbreak{}manifest.\allowbreak{}json}) before confirmatory runs. \textbf{Tier
2 --- expert-annotated texts}, of two methodologically distinct kinds
that we keep separate throughout and never pool into a single statistic,
because they test different competences. (i) \emph{Linguistic
annotation}: four clinical transcripts on which a psychoanalyst marks
discourse-relevant signifiers and rupture regions, a ground truth about
linguistic and discourse organization, scored threshold-free as the rank
AUC of channel values on marked vs.~unmarked tokens (ordered-flow term
matching). (ii) \emph{Structural/data annotation}: four pipeline
execution traces (tabular records) for which a domain expert declares
the columns a domain reader consults, a ground truth about
\emph{structural/data} organization, \emph{not} about linguistic
understanding in a direct sense. Because the mark is a whole column
rather than a term list, it is scored per column: the pooled
marked-vs-unmarked AUC is uninformative here, since the interest set
spans roughly half the tokens and is dominated by one large field. Both
kinds are fixed before any model output is seen; the clinical
annotations predate the instrument itself by roughly three quarters of a
year, and their delivery is documented. Agreement statistics are
meaningful for the structural tier, where marking a column is a
classification; for the clinical tier we argue that inter-annotator
agreement would miss the object, and validation is per reading
(Discussion, ``Ground truth without agreement'').

One text (\texttt{WFA\ Text\ 1}) served as the \emph{discovery text}:
every test below was developed and its constants fixed there. All other
texts are confirmatory; constants are not retuned per text, and results
are reported as replication counts across text \ensuremath{\times} model
\ensuremath{\times} view cells rather than as per-cell significance. The
battery (\texttt{compare\_\allowbreak{}runs.\allowbreak{}py}, run over every completed extraction)
comprises:

\begin{enumerate}
\def\labelenumi{\arabic{enumi}.}
\tightlist
\item
  \textbf{Boundary test.} Strong-anchor density per channel before vs.
  after the engineered boundary; reported as the density ratio against
  the manifest's pre-registered direction. Both views.
\item
  \textbf{Two-view correlation.} Spearman correlation of each channel's
  raw values between the operative and quiet view.
\item
  \textbf{Pin identity.} Classification of the top-decile aggregator
  anchors into \emph{delimiter} (no alphanumeric content),
  \emph{recurring lexeme} (normalized token text occurring
  \ensuremath{\geq} 5 times in the text), or \emph{other}.
\item
  \textbf{Paragraph-binding test.} Aggregator-strong anchors overlapping
  a paragraph's closing characters (identified in the operative view)
  are tested for whether they bind their own preceding paragraph: the
  mean final-layer cosine (in the anchor's subspace) of in-paragraph
  tokens minus that of all other tokens, per pin; an exact two-sided
  sign test aggregates over pins. Evaluated in both views. Paragraphs
  contribute only with \ensuremath{\geq} 5 preceding pool tokens.
\item
  \textbf{Minimal pairs.} The full A/B protocol above, run for every
  manifest-declared pair whose two runs are complete for a model.
\item
  \textbf{Tier-2 annotation scoring.} For texts with an expert
  annotation file
  (\texttt{annotations/\allowbreak{}\textless{}text\textgreater{}\allowbreak{}.\allowbreak{}json}; terms listed
  in order of appearance and matched sequentially, so repeated words
  resolve to the occurrence the expert marked; disruption passages as
  start/end markers): the rank AUC of each channel's raw values on
  annotated vs.~unannotated tokens, per view, and strong-anchor density
  ratios inside each disruption region. Annotation files record
  annotator provenance and are fixed before scoring; items outside the
  analyzed excerpt are reported and skipped. This frozen test covers the
  \emph{linguistic} annotation type (Tier 2 i). The
  \emph{structural/column} type (Tier 2 ii) carries a different
  ground-truth shape and is scored by an auxiliary per-column AUC over
  the same raw channel values, reported in Results §7, not as a frozen
  battery cell.
\item
  \textbf{Type-essence transfer.} For every pair of texts analyzed with
  the same model (manifest-declared minimal pairs excluded), word types
  present in both texts are auto-discovered (alphabetic, \ensuremath{\geq}
  5 characters, top 30 by minimum cross-text frequency, \ensuremath{\leq}
  6 occurrences per text). Each type's cross-text anchor pairs are
  scored by core-subspace overlap (\textbar A\ensuremath{\cap}B\textbar{}
  / min(\textbar A\textbar, \textbar B\textbar)) against a seeded random
  cross-text anchor baseline of 150 pairs; the baseline measures the
  trained scaffold that \emph{any} two anchors share. Because the raw
  coefficient saturates where the substrate dominates, the same
  comparison is repeated on \emph{idiosyncratic} subspaces (each text's
  dims used by \textgreater{} 50\% of its anchors removed). Same-type
  transfer at the baseline percentile means the type carries nothing
  beyond the scaffold: the structure the channels read is a property of
  the individual text, and pooling across texts by lexical type would
  average away precisely the object of study. A scope condition follows
  from corpus composition: shared types occur almost exclusively within
  register blocks (one residual cross-register pair contributes 2 of the
  391 types per model), so the reported transfer is within-register
  transfer, the case most favorable to a type essence; cross-register
  transfer is untested (see Limitations).
\end{enumerate}

\subsection{Reproducibility}\label{reproducibility}

All code is public in the
\url{https://github.com/HermeneuticAI/MetaphorTracer} repository. Every
analysis beyond the two model passes (state extraction; horizon logits)
consumes cached artifacts and is deterministic and re-runnable in
minutes on CPU. All per-text figures regenerate from the released texts
with two commands per model (extract, scan); the corpus-level results
matrix regenerates with one (\texttt{compare\_\allowbreak{}runs.\allowbreak{}py}). Every
quantitative claim in the paper is additionally recomputed from the same
artifacts by \texttt{paper\_\allowbreak{}stats.\allowbreak{}py}, which states, per claim, the
frozen value, the live value, and the source file and formula; the same
table is exposed in the app. Run folders pin the producing model; scan
artifacts carry a definition tag (\texttt{aggregator\_\allowbreak{}def}) so that
quantities computed under superseded definitions are detected and
recomputed rather than silently reused. The scan and app defaults
(\ensuremath{\mu} = 0.7, \ensuremath{\tau} = 0.7) are the ones used
throughout; legacy orchestration scripts in the repository retain
historical defaults and produce no reported figure.

\section{Results}\label{results}

\textbf{Corpus at time of writing:} 13 texts \ensuremath{\times} 3 models
(phi-4, Qwen3-8B, Llama-3.1-8B base) = 39 runs, 78
text\ensuremath{\times}model\ensuremath{\times}view cells, one engineered
minimal pair plus one discourse-framing pair (§7 follow-up; its three
runs sit alongside the corpus and enter no pooled statistic), 391 shared
word types per model for the transfer test, 4 clinical transcripts with
independent psychoanalytic annotation (Tier 2, linguistic) and 4
pipeline execution traces with a domain expert's column-of-interest
declaration (Tier 2, structural).

\subsection{1. Engineered boundaries (Tier
1)}\label{engineered-boundaries-tier-1}

\begin{figure}[H]
\centering
\pandocbounded{\includegraphics[keepaspectratio,alt={Figure 2 --- heatmap gallery}]{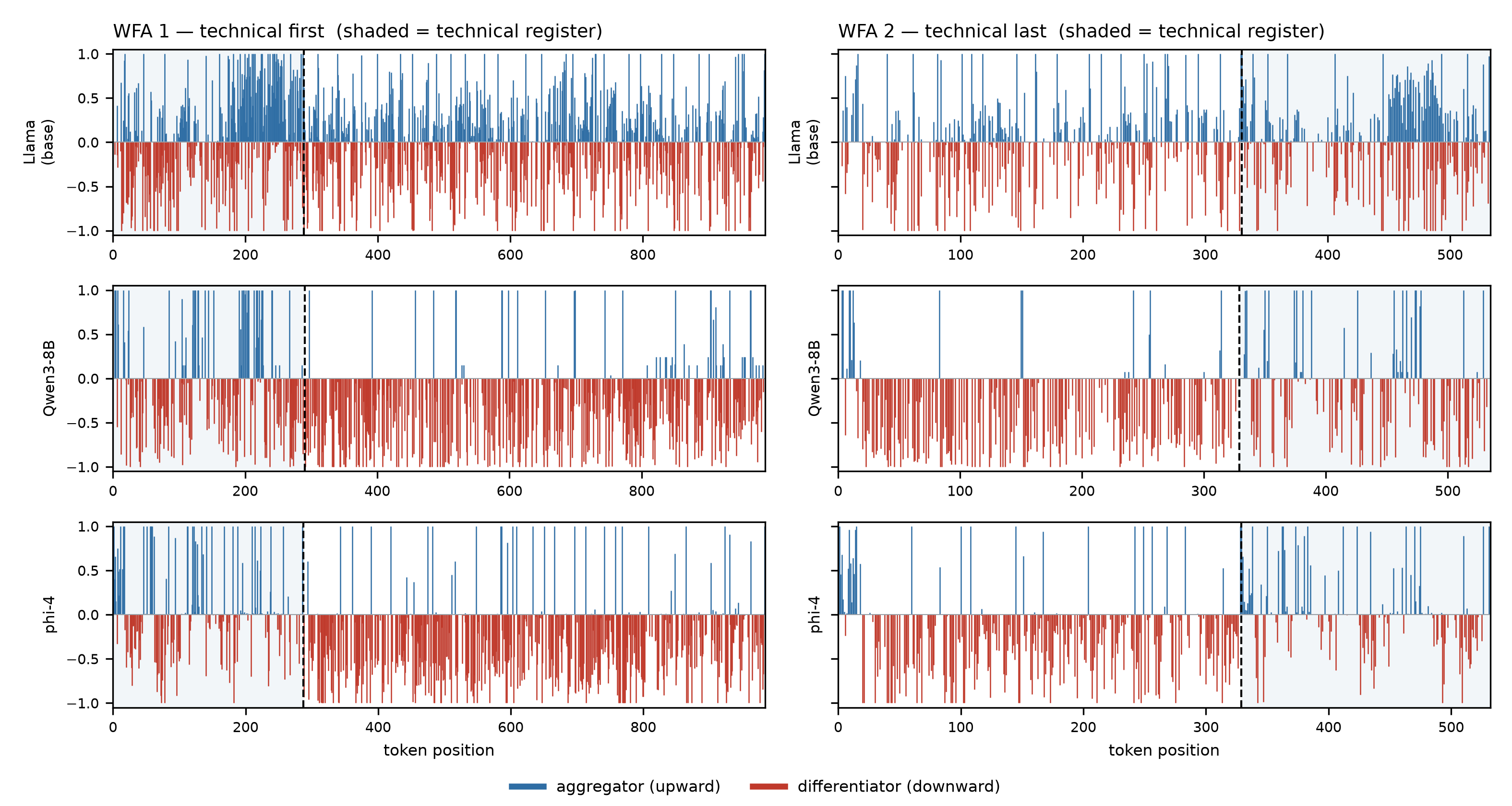}}
\end{figure}

\textbf{Figure 2 --- the aggregator concentrates in the technical
register regardless of its position.} Two-channel token maps (aggregator
upward in blue, differentiator downward in red; contrast-normalized per
run, median \ensuremath{\to} 0, 95th percentile \ensuremath{\to} 1) for
both WFA texts under all three models, the construction-known boundary
dashed, the technical register shaded. The blue mass sits in the shaded
register on both sides of the document, in categorially different token
material across models (§3).

The first battery test reads strong-anchor density across the
construction-known boundaries. The two WFA texts place the technical
register at opposite ends, deliberately so, since a positional confound
then predicts the same direction in both texts, while the register
hypothesis predicts opposite directions. Blue follows the register in
6/6 model \ensuremath{\times} text cells:

{\def\LTcaptype{none} 
\begin{longtable}[]{@{}llll@{}}
\toprule\noalign{}
text (prediction) & Llama-3.1-8B & Qwen3-8B & phi-4 \\
\midrule\noalign{}
\endhead
\bottomrule\noalign{}
\endlastfoot
WFA 1, technical first (blue before) & 2.57 & 2.99 & 3.04 \\
WFA 2, technical last (blue after) & 0.66 & 0.39 & 0.57 \\
\end{longtable}
}

The differentiator (red) channel concentrates in the soft/marketing
register in the instruction-tuned models (WFA 1: 0.43, 0.16; WFA 2:
1.63, 4.14) but not in the base model (1.21, 0.97), the first of several
instruct/base dissociations below, and one the matched-lineage follow-up
(Methods; §6) assigns to tuning in the strict sense: the base's instruct
twin concentrates like the tuned models (0.33; 3.54). Stated for
completeness, the twin's aggregator blurs the reversed-order control
(before-ratio 1.05 against the base's 0.66), the one follow-up readout
on the wrong side of 1; it enters no pooled count. The hard-jargon
control text (no pre-registered direction) splits the models (blue
before-ratio 0.41--3.10), consistent with opaque text being where
reader-dependence is largest.

\subsection{2. Two views, two maps}\label{two-views-two-maps}

\begin{figure}[H]
\centering
\pandocbounded{\includegraphics[keepaspectratio,alt={Figure 3 --- two-view boundary inversion}]{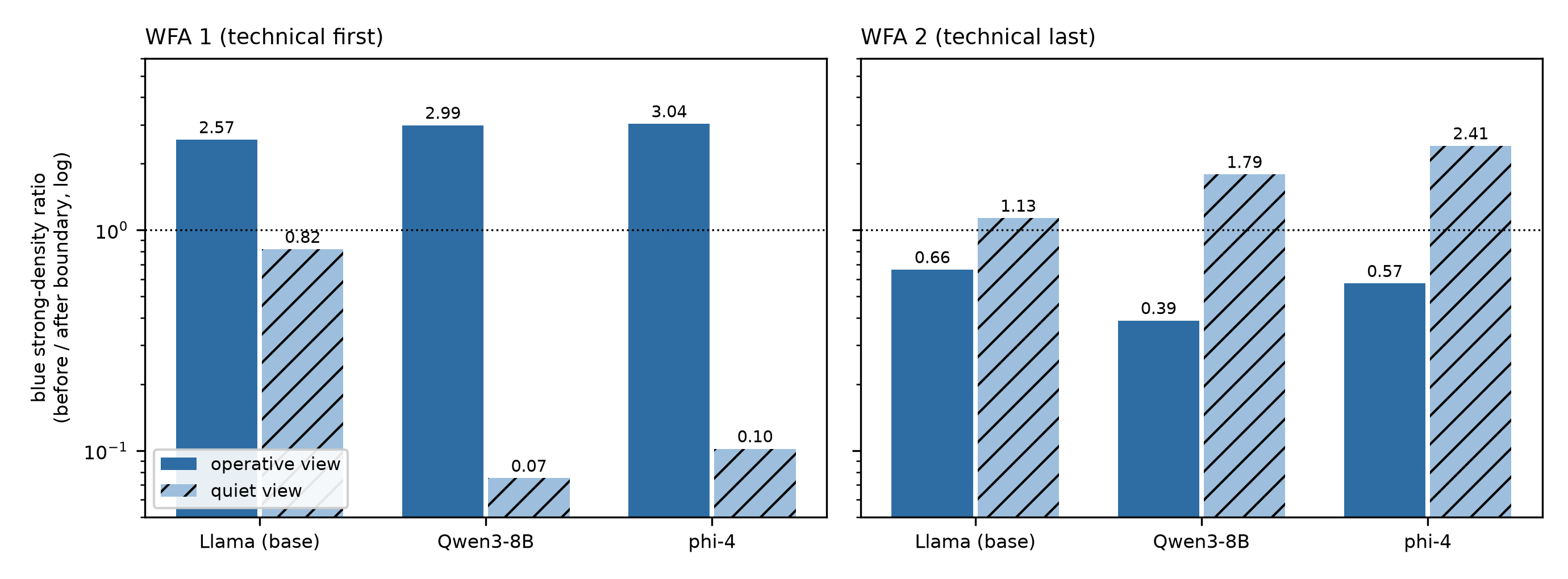}}
\end{figure}

\textbf{Figure 3 --- the two-view boundary inversion.} Aggregator
strong-density ratio (before/after the construction-known boundary, log
scale) for both WFA texts \ensuremath{\times} three models, operative
against quiet view. The ratio crosses the 1-line between views in both
boundary directions: the operative view tracks the technical register,
the quiet view the soft register.

Rescanning in the quiet-dimensions view (per-layer z-scoring) does not
weaken the boundary alignment, it \emph{inverts} it, in all three models
and in both boundary directions: strongly in the instruction-tuned
models (quiet blue on WFA 1: 0.08, 0.10 against operative
3.0\ensuremath{\times}; WFA 2 reversed design: 1.79, 2.41), weakly but on
the inverted side in the base model (0.82 and 1.13). Per-anchor channel
values correlate weakly or negatively between views
(e.g.~\ensuremath{-}0.23, blue, phi-4, WFA 1). The two geometries are
not noisy variants of one map; they are two maps.

\subsection{3. Pin identity}\label{pin-identity}

\begin{figure}[H]
\centering
\pandocbounded{\includegraphics[keepaspectratio,alt={Figure 4 --- channel exclusion and pin depth}]{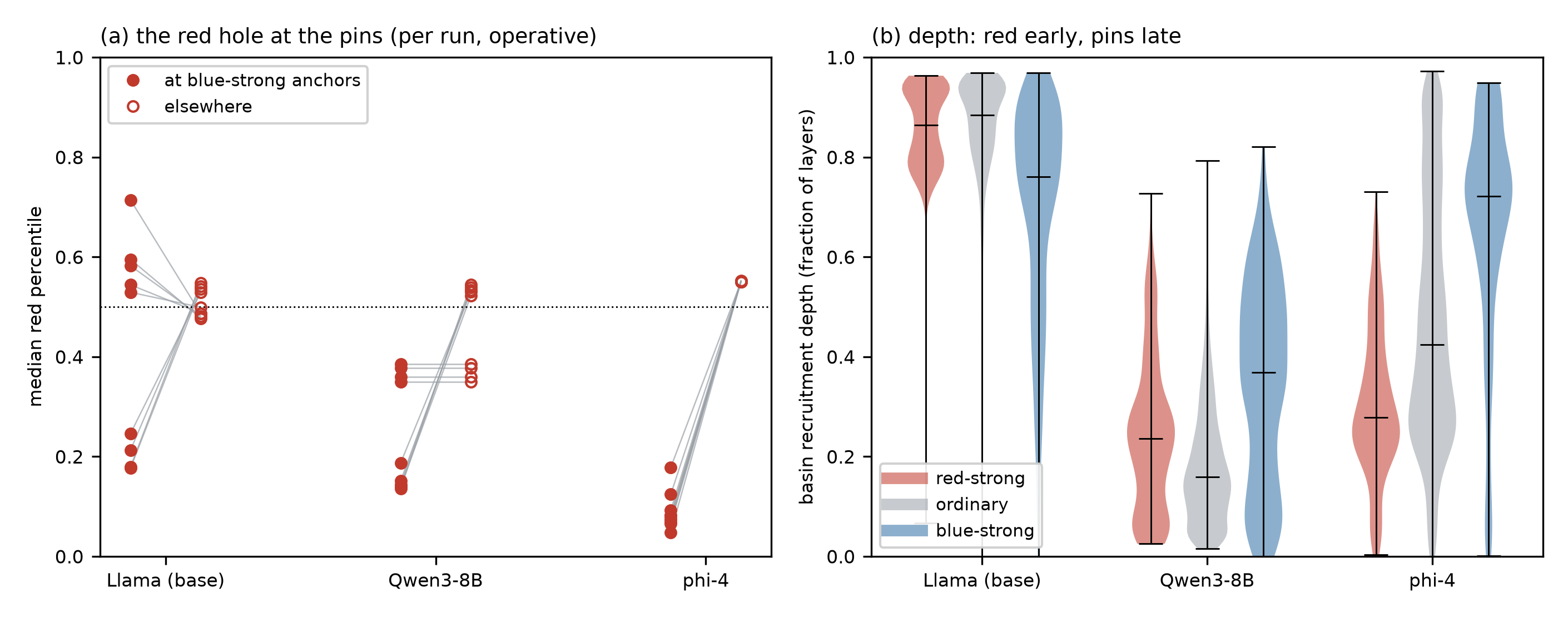}}
\end{figure}

\textbf{Figure 4 --- channel exclusion and recruitment depth.} (a)
Median red-channel percentile at blue-strong anchors against elsewhere,
per run (operative view, the nine non-trace texts): the red hole at the
pins in the tuned models; in the base model the engineered runs show no
hole and the clinical runs a partial one. Post-hoc, descriptive;
confound checked in the text. (b) Anchor-level basin recruitment depth
pooled over the same runs: red-strong early, blue-strong late, ordinary
anchors between, the pins bind on the way out of the reported workspace
regime (Discussion).

Battery test 3 asks what the pins are made of, and the answer differs by
model: \emph{delimiters} carry the top of the ranking in phi-4 (18/20
top anchors on the discovery text, 19--20/20 on the clinical
transcripts, 6--14/20 on the remaining engineered texts) and in the base
model (14--20/20 throughout), and are absent from Qwen3-8B's (0--2/20
across all nine texts): its pin material is lexical, the text's
recurring lexemes and content words.

These are counts, and the warning belongs where the aggregate is used: a
count cannot express rank structure. In phi-4 the delimiters form a
crust over a content body: ranks 1--20 of a clinical run's top decile
are 95--100\% delimiters, the decile's remainder 63--73\%, and the
content anchors sit at median ranks more than twice as deep. The base
model, by contrast, interleaves the two strata (crust 0.70--0.85, the
strata's median ranks close or inverted). What the crust masks is shown
in §8: stripped of every delimiter, the alignment with the analyst's
marks rises. Three follow-up measurements, and one post-hoc observation:

\begin{itemize}
\tightlist
\item
  \textbf{Paragraph binding (capiton test).} phi-4's paragraph-final
  pins bind their own preceding paragraph above the rest of the text: on
  the discovery text, operative \ensuremath{\Delta}cos = +0.022, 14/18 pins
  positive (sign test p = 0.031); quiet \ensuremath{\Delta}cos = +0.066,
  15/18, p = 0.0075, and the direction replicates on the four other
  texts with eligible pins. Qwen produces no paragraph-final pins (its
  pin material is lexical), so the test does not apply; the material
  difference is a result, not a missing value.
\item
  \textbf{Depth.} Pin basins saturate late: median recruitment layer
  29/40 (\ensuremath{\approx} 72\% of depth) vs 17/40 (\ensuremath{\approx} 42\%)
  for ordinary anchors and 11/40 (\ensuremath{\approx} 28\%) for red-strong
  ones. The pins bind the text where reported global-workspace content
  collapses toward output representations, not where it ignites.
\item
  \textbf{Type transfer} (§6) shows the pin function does not travel
  with the token type across texts in the tuned models.
\item
  \textbf{Channel exclusion (post-hoc; found after freeze,
  descriptive).} In the tuned models' operative view the two channels
  are mutually exclusive at the top: the top-decile blue and red sets
  are disjoint in 9/9 (phi-4) and 8/9 (Qwen3-8B) runs, and the median
  red percentile at blue-strong anchors is 0.05--0.18 in phi-4 against
  \ensuremath{\approx}0.55 elsewhere; the pins sit in a red hole. In Qwen's
  clinical runs the exclusion is near-exact (the majority of blue-strong
  anchors carry red of literally zero). The base model shows no hole on
  the engineered texts (red at blue-strong at the 0.53--0.71 percentile,
  strong-set overlap up to 0.167) and a partial hole on the clinical
  ones (0.18--0.25), a further instruct/base dissociation, and one the
  matched twin (§6) splits down the middle: under tuning the hole opens
  everywhere (red at blue-strong at the 0.13--0.36 percentile on the
  engineered texts, 0.14--0.20 on the clinical, against 0.54 elsewhere),
  while the recruitment order does not move, blue-strong later than
  red-strong in 1/9 runs, exactly the base's count. The exclusion is
  tuning-borne; the depth ordering is lineage. The quiet view
  \emph{inverts} the exclusion (red at blue-strong anchors at the 0.66
  percentile vs 0.48 elsewhere). Depth separates the strong sets the
  same way: phi-4's blue-strong basins recruit later than its red-strong
  basins in 9/9 runs (0.52--0.77 vs 0.21--0.42 fractional depth; Qwen
  7/9; the base model 1/9, inverted). A confound has to be stated here:
  blue is rank-coupled to subspace size k, and in phi-4 pull\_frac\_gap
  anti-correlates with k (median \ensuremath{-}0.42), so part of phi's
  hole could be arithmetic compression of the gap in large subspaces,
  but Qwen shows the full exclusion at near-zero k-coupling
  (\ensuremath{-}0.09), and the base model's large-k anchors show no
  hole (\ensuremath{-}0.07), so the exclusion is not an arithmetic
  necessity of large subspaces.
\end{itemize}

\subsection{4. The engineered anomaly (minimal
pair)}\label{the-engineered-anomaly-minimal-pair}

\begin{figure}[H]
\centering
\pandocbounded{\includegraphics[keepaspectratio,alt={Figure 5 --- suffix divergence}]{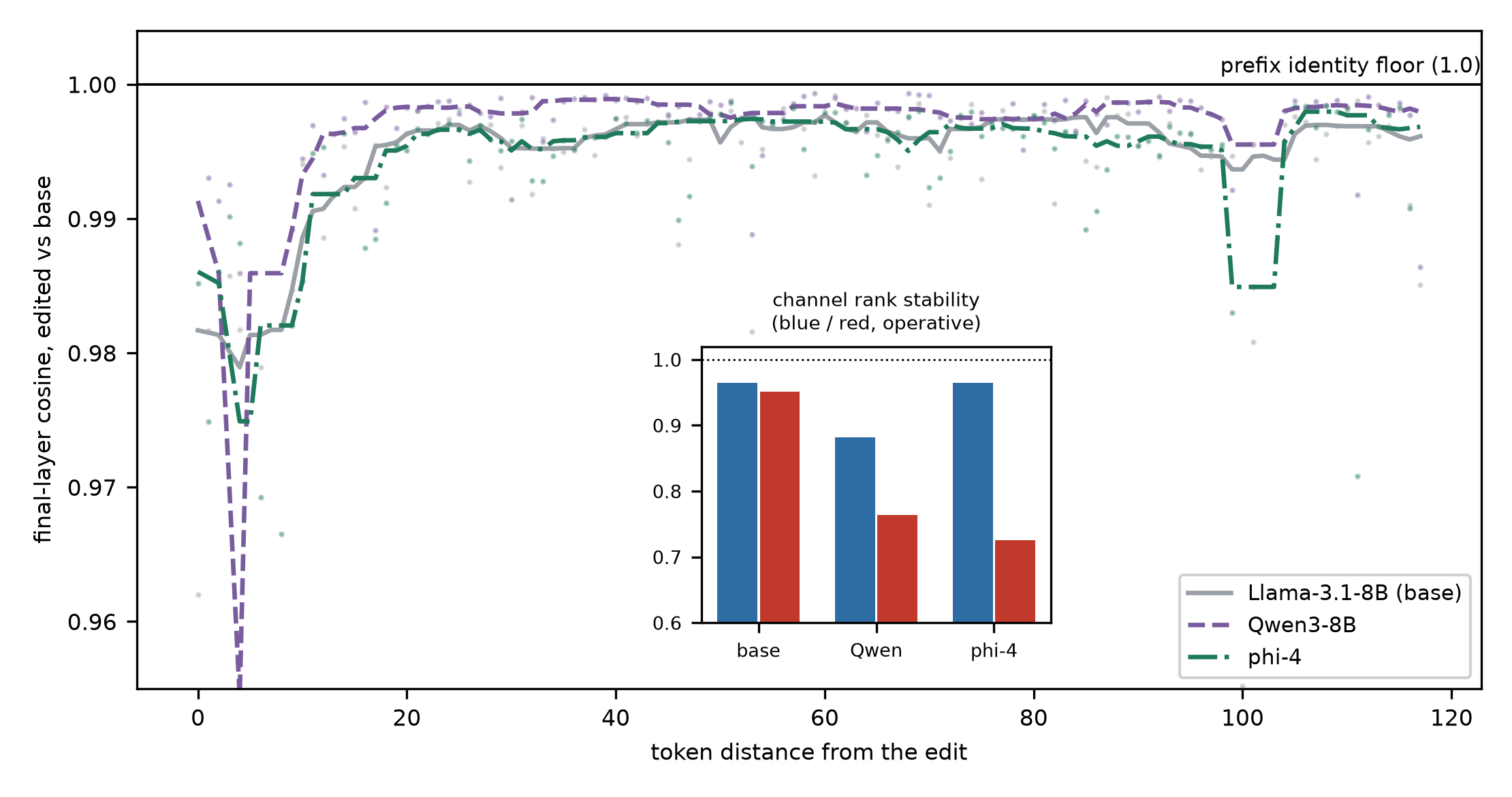}}
\end{figure}

\textbf{Figure 5 --- suffix divergence under the engineered anomaly.}
Per-token final-layer cosine between the edited and base versions over
the shared suffix, by distance from the edit (dots: tokens; lines:
rolling median), with the prefix identity floor at 1.0. States diverge
near the edit and recover with distance, never to identity. Inset:
channel rank stability over the same aligned tokens, high, but
systematically below state stability, with the largest drops in the
differentiator of the tuned models.

Battery test 5 sets the absurd passage (deliberately false information,
\textasciitilde110 tokens) against its excised base version:

\begin{itemize}
\tightlist
\item
  \textbf{Causality floor.} Prefix states are identical between
  versions: minimum final-layer cosine 1.0 (float32) over the 741--742
  prefix tokens (tokenizer-dependent), all models.
\item
  \textbf{Region signature.} In the edited text, the anomaly is an
  aggregation void with elevated operative differentiation (blue
  strong-density 0.08--1.0\ensuremath{\times}, red
  1.27--2.43\ensuremath{\times}) and quiet-view silence. At a
  token-count-matched window at the base text's junction the signature
  is absent (red 0.77--1.00\ensuremath{\times} in 3/3 models): it belongs to
  the content, not the position.
\item
  \textbf{What a mundane reader sees here
  (\texttt{baseline\_\allowbreak{}regions.\allowbreak{}py}).} Scored as ``which tokens lie inside
  the edit'', the predictive measures and the aggregator separate in
  sign: entropy 0.63--0.67 and word surprisal 0.63--0.65 against blue
  0.17--0.32, all three models. The lexical predictors are indifferent
  (rarity 0.53--0.54, length and content 0.49--0.50): the passage is
  absurd, not unusual. Blue's own junction control holds (0.42--0.61).
  Where a passage is maximally improbable and woven into nothing, the
  predictive measures rise and the channel leaves.
\item
  \textbf{Suffix contamination.} Identical wording read after the
  anomaly diverges from its base-version states: median cosine
  0.987--0.993 near the edit, recovering to 0.996--0.998 at distance,
  never to identity.
\item
  \textbf{Channel stability.} Over the shared suffix, channel rank
  stability is high but systematically below state stability: Spearman
  0.73--0.97, with the largest drops in the differentiator channel of
  the tuned models (0.73, 0.76). Structural value is roughly an order of
  magnitude more edit-sensitive than the states it is read from.
\end{itemize}

\subsection{5. Clinical transcripts (Tier
2)}\label{clinical-transcripts-tier-2}

\begin{figure}[H]
\centering
\pandocbounded{\includegraphics[keepaspectratio,alt={Figure 6 --- Tier-2 AUC forest}]{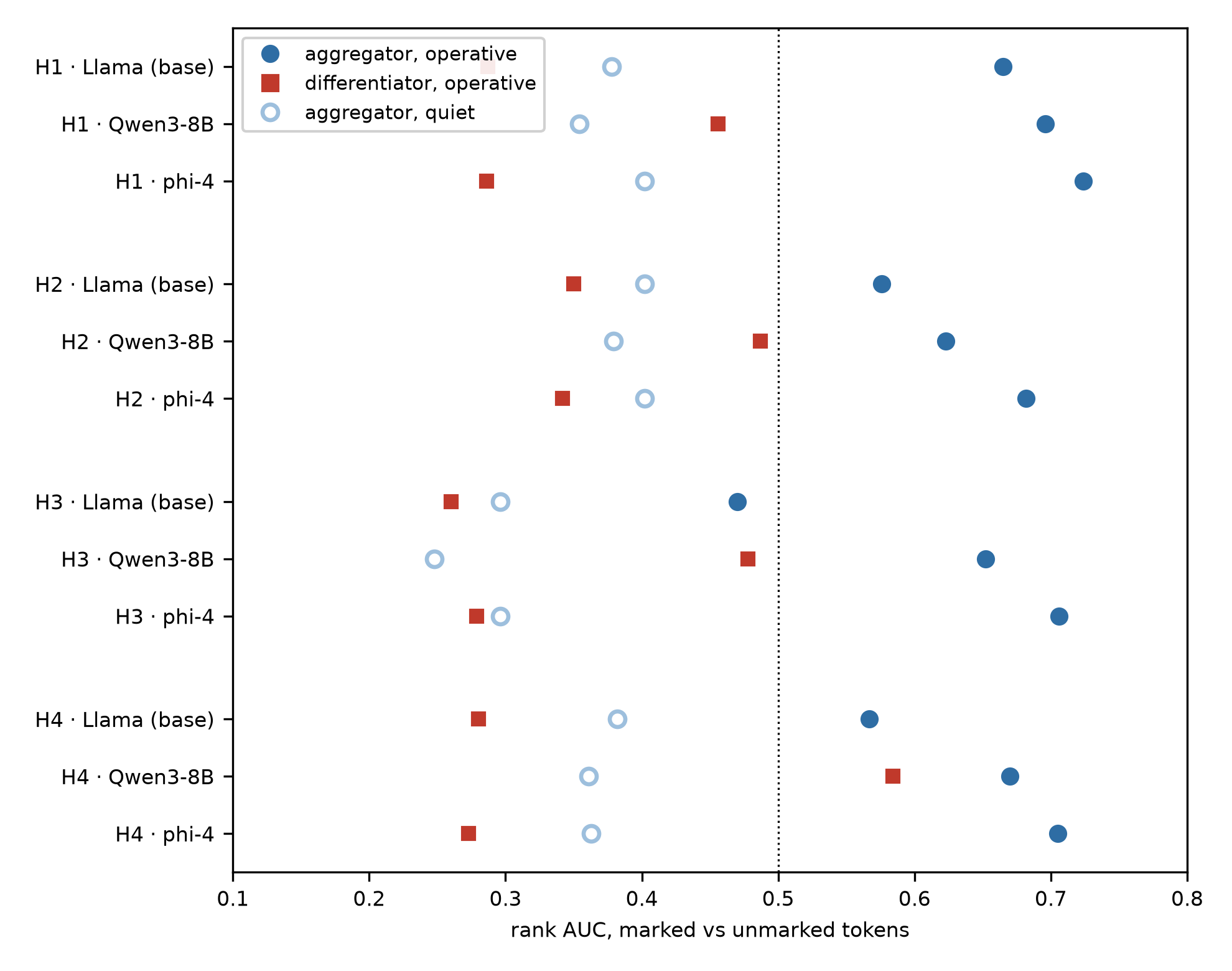}}
\end{figure}

\textbf{Figure 6 --- Tier-2 annotation scoring.} Rank AUC on
analyst-marked against unmarked tokens for the three channels of
interest, four transcripts \ensuremath{\times} three models, chance at 0.5.
Blue-operative above chance in all but one cell (base model, transcript
3); red-operative at or below chance in all but one (Qwen3-8B,
transcript 4); quiet-blue below chance throughout, the pre-registered
pattern, 34/36 directional cells. (\texttt{fig\_\allowbreak{}tier2\_\allowbreak{}forest.\allowbreak{}py})

Battery test 6 scores the independent psychoanalytic annotation (four
transcripts, 12--29 marked items each), threshold-free. The directional
prediction was registered after transcript 1: blue-operative AUC above
0.5, red and quiet-blue at or below. The outcome is 34/36 directional
cells; since the direction was fixed on that first transcript, its nine
cells are in sample, and the three later transcripts hold in 25 of their
27. The count is a replication tally and not thirty-six independent
tests: the three predictions per transcript are made of channels the
paper elsewhere shows to be related, blue and red being mutually
exclusive at the top of the ranking (§3) and the two views inverting on
the engineered boundary (§2), so the red and quiet-blue predictions are
the weaker half of the count. We therefore report the three components
separately below, and the alignment claim rests on the blue-operative
column.

{\def\LTcaptype{none} 
\begin{longtable}[]{@{}llll@{}}
\toprule\noalign{}
transcript & Llama blue-op & Qwen blue-op & phi-4 blue-op \\
\midrule\noalign{}
\endhead
\bottomrule\noalign{}
\endlastfoot
H1 & 0.665 & 0.696 & 0.724 \\
H2 & 0.576 & 0.623 & 0.682 \\
H3 & 0.470 \ensuremath{\times} & 0.652 & 0.706 \\
H4 & 0.567 & 0.670 & 0.705 \\
\end{longtable}
}

The fourth transcript replicates both the direction and the
per-transcript model ordering (Llama \textless{} Qwen \textless{}
phi-4). Red-operative is below chance in 11/12 cells (0.26--0.49; the
miss is Qwen3-8B on H4 at 0.58); quiet-blue below chance in 12/12
(0.25--0.40). The two misses are the base model's H3 blue (the weakest
tracker throughout, §6) and one tuned red cell. The matched instruct
twin (§6; never pooled) adds twelve cells in the registered directions,
12/12: blue-operative above chance on every transcript (0.570--0.702,
median 0.650 against the base's 0.571), red-operative below
(0.26--0.35), quiet-blue below (0.32--0.45); the base's one blue miss,
H3, crosses chance at 0.570.

\textbf{What the annotation can settle.} The marks are a probe of the
channel, not its criterion. What they make available is the structure a
Lacanian reading would assume, the points at which a discourse is
organized, and the question is whether the channel is deposited there.
They never exhaust the strong-aggregation set: in phi-4 and the base
model roughly 60--75\% of strong aggregation sits on bare delimiters,
and further strata remain once the marked terms are removed (Discussion,
``The unmarked lattice''). A high AUC would therefore not mean that the
channel is the annotation, and a missed mark does not mean the channel
is empty there.

Static lexical baselines were run on the same split (Methods). Rarity
recovers the marked terms at pooled AUC 0.875--0.878, invariant across
the three readers to 0.02, and it follows the channel nowhere else in
the paper: not at the engineered anomaly, where it is indifferent while
blue is signed and absent (§4), and not at the marked disruptions, where
it sits at the floor while blue is extreme (below). A type-level
constant, one number per word and the same table for every reader,
cannot follow results that turn on occurrence, position and grain. It
aligns at one grain of one ground truth, so blue is reported under the
controls that grain requires: above chance within content words in 10/12
cells (0.53--0.67) and residualized on word-level surprisal and rarity
in 11/12 (0.52--0.66). Pooled, and in each case raw, then within content
words, then residualized, the base model runs at 0.562, 0.549 and 0.585,
Qwen3-8B at 0.657, 0.541 and 0.518, and phi-4 at 0.703, 0.602 and 0.637.
Within content words the two misses are the base model's H3, its known
miss, at 0.466 and Qwen3-8B's H2 at 0.497; residualized, only Qwen3-8B's
H2 stays below chance, at 0.495. Against the two permutation nulls phi-4
clears both (p = 0.0001 block, 0.0013 rarity-matched), the base model
the block null only (0.0003 against 0.103), and Qwen3-8B neither (0.21,
0.47). The readers' pooled spread falls from 0.141 to 0.119 under the
subtraction, and to 0.052 between the two the method adjudicates
(per-cell values in \texttt{outputs/\allowbreak{}comparison/\allowbreak{}baseline\_\allowbreak{}aucs.\allowbreak{}csv}).

\textbf{The readers are routed through different material.} Why the
subtraction lands so differently by reader is itself measurable, and
measurable without the marks. Over every scanned token, the share of
blue's rank variance the two covariates explain is 0.15--0.22 in
Qwen3-8B, 0.04--0.08 in phi-4 and 0.00--0.03 in the base model, disjoint
across all four transcripts; since the annotation never enters the
quantity, it conditions the subtraction without being a function of its
outcome. This is the continuous form of the pin-material dissociation of
§3, and it resolves the section's asymmetries into one structure.
Qwen3-8B's ranking is lexical throughout, so the subtraction removes a
fifth of the channel itself, and on that reader's clinical increment the
method decides in neither direction; we claim nothing there. The two
delimiter-crusted readers are layered instead: their organizing
positions are the most frequent material in the text, which is why
erasing delimiters raises phi-4's alignment rather than lowering it
(§8), while the content anchors beneath the crust overlap the rare
stratum the baselines measure, which is why the subtraction still costs
phi-4 0.066. The crust is neither rare nor marked, and the reading the
annotation scores happens underneath it.

\textbf{The prediction measures anti-align.} Predictive entropy ranks
marked tokens at 0.30--0.32 pooled and realized token surprisal at
0.43--0.45, both below chance. Token surprisal below chance is partly
mechanical (long compounds carry near-deterministic continuation tokens;
projecting each word's maximum token surprisal back over its tokens
repairs the baseline to 0.69), but the repaired baseline collapses
exactly where the channel holds: under repetition. Across the three
marked repetitions of Wahrheit (H3), the word's surprisal percentile
falls from 0.49--0.57 to 0.33--0.55 to 0.09--0.16 across the three
models while its blue percentile stays at 0.73--0.93; Mutter, twice
marked in H2, sits at surprisal 0.02--0.07 with blue at 0.82--0.93; the
second marked Selbstmord at 0.08--0.20 against blue 0.80--1.00;
gestorben (H1) at 0.10--0.20 against 0.87--0.98. The pattern is not
universal (the second Generation falls on both, blue 0.32--0.52,
surprisal 0.11--0.13), but its direction is structural: repetition
destroys information content, not markedness, and a predictability
reader loses the signifier at the moment it begins to insist
(Discussion, ``Structural value''; per-term percentiles in
\texttt{outputs/\allowbreak{}comparison/\allowbreak{}baseline\_\allowbreak{}terms.\allowbreak{}csv}).

The exemplars are not selected: the table gives every marked term with
two or more occurrences in its transcript (11 terms), the cross-model
range of the word's surprisal, attention-rollout and blue percentile by
occurrence (\texttt{baseline\_\allowbreak{}repetition.\allowbreak{}csv},
\texttt{attention\_\allowbreak{}repetition.\allowbreak{}csv}; attention rollout is the
model-internal rival instrument, scored in full in §8). Across all 33
term \ensuremath{\times} model cases the surprisal percentile falls from
first to last occurrence in 26, with a median change of
\ensuremath{-}0.16; the rollout percentile falls in 32, with a median
change of \ensuremath{-}0.18, the steepest drain of the three; blue
falls in 21, with a median change of \ensuremath{-}0.02. Under
repetition the information measure and the traffic measure drain
together, and the aggregator holds.

{\def\LTcaptype{none} 
\begin{longtable}[]{@{}
  >{\raggedright\arraybackslash}p{(\linewidth - 6\tabcolsep) * \real{0.2500}}
  >{\raggedright\arraybackslash}p{(\linewidth - 6\tabcolsep) * \real{0.2500}}
  >{\raggedright\arraybackslash}p{(\linewidth - 6\tabcolsep) * \real{0.2500}}
  >{\raggedright\arraybackslash}p{(\linewidth - 6\tabcolsep) * \real{0.2500}}@{}}
\toprule\noalign{}
\begin{minipage}[b]{\linewidth}\raggedright
term
\end{minipage} & \begin{minipage}[b]{\linewidth}\raggedright
surprisal by occurrence
\end{minipage} & \begin{minipage}[b]{\linewidth}\raggedright
rollout by occurrence
\end{minipage} & \begin{minipage}[b]{\linewidth}\raggedright
blue by occurrence
\end{minipage} \\
\midrule\noalign{}
\endhead
\bottomrule\noalign{}
\endlastfoot
Selbstaufgabe (H1) & 0.91--0.97 \ensuremath{\to} 0.69--0.80 & 0.29--0.56
\ensuremath{\to} 0.03--0.07 & 0.90--0.98 \ensuremath{\to} 0.80--0.97 \\
Generation (H2) & 0.58--0.81 \ensuremath{\to} 0.11--0.13 & 0.87--0.92
\ensuremath{\to} 0.54--0.76 & 0.62--0.81 \ensuremath{\to} 0.32--0.52 \\
Mutter (H2) & 0.02--0.07 \ensuremath{\to} 0.06--0.07 & 0.86--0.87
\ensuremath{\to} 0.75--0.86 & 0.82--0.93 \ensuremath{\to} 0.84--0.90 \\
Selbstmord (H2) & 0.67--0.98 \ensuremath{\to} 0.08--0.20 & 0.89--0.93
\ensuremath{\to} 0.81--0.84 & 0.84--0.95 \ensuremath{\to} 0.80--1.00 \\
Werbung (H2) & 0.46--0.68 \ensuremath{\to} 0.68--0.83 & 0.34--0.63
\ensuremath{\to} 0.22--0.27 & 0.56--0.90 \ensuremath{\to} 0.60--0.90 \\
Überleben (H2) & 0.84--0.96 \ensuremath{\to} 0.32--0.47 & 0.53--0.62
\ensuremath{\to} 0.29--0.64 & 0.81--0.95 \ensuremath{\to} 0.70--0.93 \\
Gutes (H3) & 0.92--0.96 \ensuremath{\to} 0.81--0.91 & 0.68--0.73
\ensuremath{\to} 0.02--0.16 & 0.60--0.94 \ensuremath{\to} 0.84--0.89 \\
Make-Up (H3) & 0.99 \ensuremath{\to} 0.55--0.89 & 0.60--0.65
\ensuremath{\to} 0.32--0.43 & 0.25--0.75 \ensuremath{\to} 0.32--0.72 \\
Wahrheit (H3) & 0.49--0.57 \ensuremath{\to} 0.33--0.55 \ensuremath{\to}
0.09--0.16 & 0.76--0.87 \ensuremath{\to} 0.70--0.73 \ensuremath{\to}
0.69--0.71 & 0.81--0.93 \ensuremath{\to} 0.80--0.93 \ensuremath{\to}
0.73--0.93 \\
Wertvolles (H3) & 0.87--0.94 \ensuremath{\to} 0.78--0.96 & 0.08--0.12
\ensuremath{\to} 0.00 & 0.13--0.76 \ensuremath{\to} 0.54--0.91 \\
Liebe (H4) & 0.20--0.79 \ensuremath{\to} 0.41--0.49 & 0.75--0.82
\ensuremath{\to} 0.27--0.64 & 0.16--0.64 \ensuremath{\to} 0.04--0.34 \\
\end{longtable}
}

Two qualitative structures sit inside the aggregate, and both replicate:

\begin{itemize}
\tightlist
\item
  \textbf{The misses are graded, the deep stratum is lexically specific,
  and the pattern replicates.} Scored by the best rank any of its tokens
  reaches (span-maximum blue percentile across models,
  \texttt{baseline\_\allowbreak{}terms.\allowbreak{}csv}), H1's marked items separate into three
  strata with clean gaps: the conflict/destruction lexicon is caught
  outright (gestorben, Raubbau, Befreiungsschlag, Selbstaufgabe,
  Weißglut, maxima 0.93--0.99), three process-and-affect terms sit just
  under the strong-decile cutoff (Aufarbeitung, hingucke, Wärme,
  0.83--0.89, edge cases of the threshold rather than declines), and the
  only deep misses are the professions of love, Liebe pur and nur Liebe,
  whose best token in any model stops at 0.53. H4, whose annotation was
  fixed in the same sitting as the others, months before the instrument
  existed, reproduces the deep stratum exactly: pure Liebe and Liebe
  twice at 0.04--0.64, Kind at 0.66, with every other marked item above
  0.92. Wärme makes the specificity sharp: warmth is nearly caught in
  all three models (0.88--0.89) while love is declined in all three; the
  decline attaches to the Liebe lexeme, not to affect vocabulary.
  Whatever the aggregator consolidates, professions of love are
  systematically not it. The mundane baselines sharpen the observation
  into a dissociation: the same spans rank high on the cheap predictors
  (span-maximum rarity percentile 0.86--0.91 on the \emph{pur} phrases,
  word-level surprisal 0.76--1.00), so the miss is no frequency
  artifact. The baselines declare these spans easy, and the aggregator
  declines them, in all three models.
\item
  \textbf{Ruptures leave channel traces --- as extremes, not as mere
  deviations.} The guiding hypothesis was that marked ruptures would be
  visible in the channels; they are, though not universally, as the
  complexity of the problem of rupture would lead one to expect. The
  claim only has content against a base rate. Deviation alone carries no
  information: 99\% of arbitrary unmarked windows of the same sizes
  deviate by \ensuremath{\geq} 1.5\ensuremath{\times} or \ensuremath{\leq}
  0.67\ensuremath{\times} in at least one of the twelve
  model\ensuremath{\times}channel\ensuremath{\times}view cells (108/109 in a
  sliding-window scan, \texttt{rupture\_\allowbreak{}windows.\allowbreak{}py}), and six of the
  seven marked disruptions do too. What carries information is rank.
  Held against every unmarked window of its own size in its transcript,
  the collapsed-syntax passage of H3 is the single most
  quiet-blue-elevated window in all three models (quiet blue
  2.2--2.9\ensuremath{\times}, though its length admits only 11 comparison
  windows), and its second marked span sits at p95--p98. H2's principal
  marked passage is the single most quiet-blue-withdrawn window in all
  three models (0.42--0.49\ensuremath{\times}), while H2's second span and
  H1's two spans are extreme in one model and middling in the others.
  H4's single marked span is the exception: one long undivided passage,
  \textasciitilde27\% of the transcript over three speaker turns, it
  deviates in no cell (all ratios 0.72--1.40) and is too long to rank,
  admitting only \textasciitilde6 comparison windows. Segmented at the
  speaker-turn boundaries, a segmentation that is ours and not the
  annotation's, its sub-spans trace in opposite directions that cancel
  (Discussion). The engineered anomaly keeps its separate control: its
  red rim over an aggregation void appears at the edit and not at a
  token-matched junction window. The same scan returns unmarked windows
  carrying the same replicated signatures, treated in the Discussion. A
  potential typology of rupture kinds is sketched there, pending a
  larger annotated corpus.
\item
  \textbf{At this grain the lexical measures invert, and one rival
  remains.} Rescoring the same windows with mundane features in place of
  the channels (\texttt{baseline\_\allowbreak{}regions.\allowbreak{}py}; same spans, same density
  ratio, same threshold, same ranking) puts content-word density at the
  floor of the distribution for every marked span in every model, and
  rarity at 0.00--0.18 on the two spans whose channel signature
  replicates: the marked disruptions are the least lexically conspicuous
  windows of their transcripts, so the same analyst marks lexically
  salient items at the term grain and lexically inconspicuous passages
  at this one. One measure does track the channel here, and it is not
  lexical: predictive entropy is the single most elevated window of its
  size in all three models on the collapsed-syntax passage, the same
  span carrying the headline quiet-blue elevation. It cannot, however,
  sign the disturbance. Entropy is elevated on H2's principal passage
  too (0.82--1.00), where the channel is maximally withdrawn, so it
  reports that something is disturbed while the channel reports in which
  direction, which is what the typology below rests on; and by the
  all-three-models criterion it replicates on that one span alone.
\end{itemize}

Group-discussion polyphony attenuates the alignment (H2, where the
channels place strong aggregation on the other speakers' discourse,
which the annotation leaves unmarked; Discussion, ``The unmarked
lattice''), consistent with the annotation tracking one speaker's
discourse.

\begin{figure}[H]
\centering
\pandocbounded{\includegraphics[keepaspectratio,alt={Figure 7 --- transfer/fidelity inversion}]{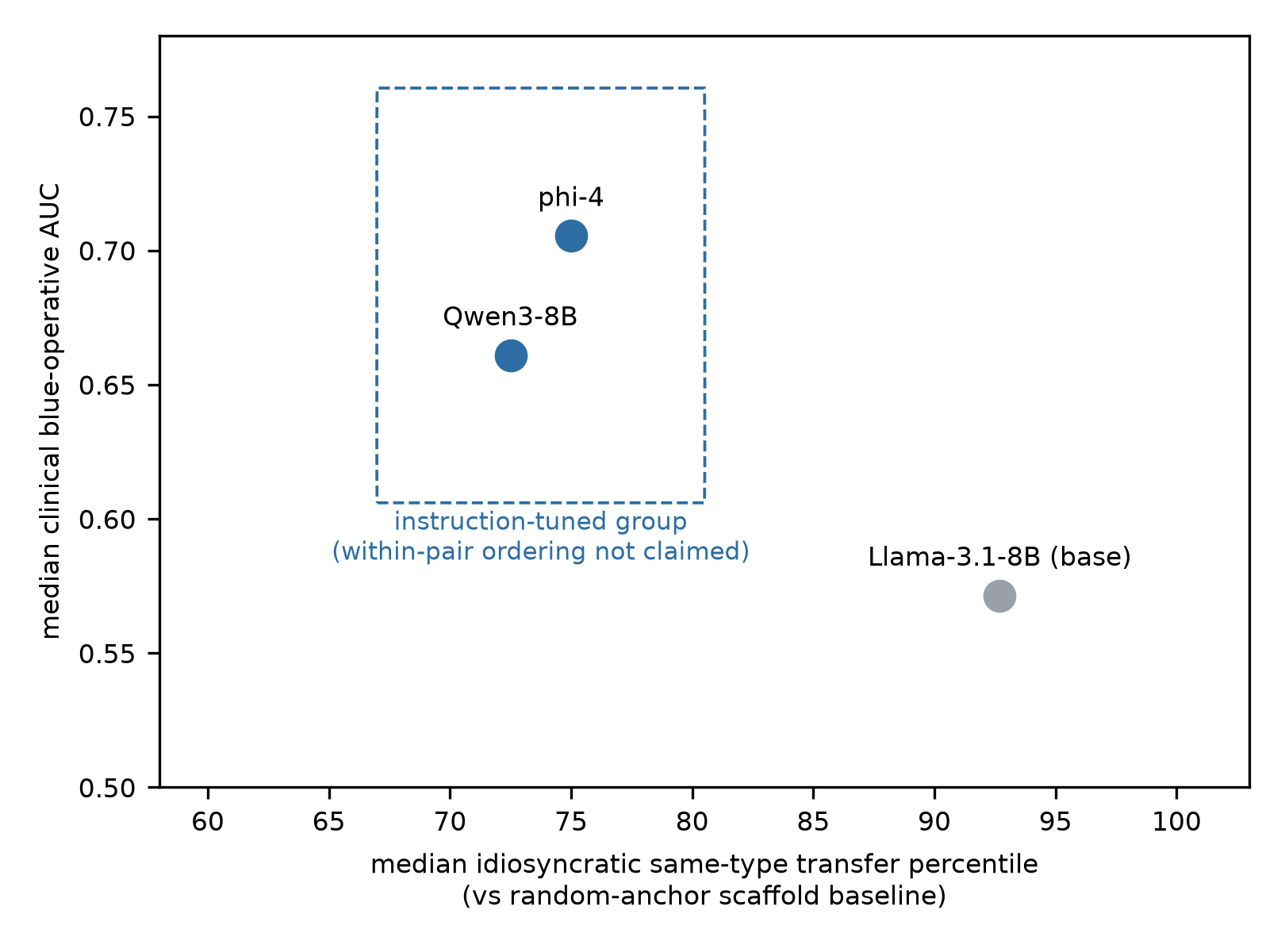}}
\end{figure}

\textbf{Figure 7 --- the transfer/fidelity inversion.} Median
idiosyncratic same-type transfer percentile (x) against median clinical
blue-operative AUC (y), per model. The type-anchored reader, the base
model, transferring near the 93rd percentile of the scaffold baseline,
reads the singular discourse worst; the tuned models transfer near
baseline and read it best. Fidelity here is raw tracking of the
annotation as a whole; the readers differ in how far their channels ride
the shared lexical plane, so the contrast is reported here raw and
graded under the controls in §5. The ordering within the tuned pair is
not stable across corpus compositions and is not claimed. The matched
instruct twin (not shown; below) breaks the axis into its two
components: base-level transfer, near-tuned fidelity.

\subsection{6. Type-essence transfer}\label{type-essence-transfer}

Battery test 7 scores same-type cross-text subspace overlap against a
random-anchor scaffold baseline, on idiosyncratic dimensions, for the
391 shared word types per model:

{\def\LTcaptype{none} 
\begin{longtable}[]{@{}
  >{\raggedright\arraybackslash}p{(\linewidth - 6\tabcolsep) * \real{0.2500}}
  >{\raggedright\arraybackslash}p{(\linewidth - 6\tabcolsep) * \real{0.2500}}
  >{\raggedright\arraybackslash}p{(\linewidth - 6\tabcolsep) * \real{0.2500}}
  >{\raggedright\arraybackslash}p{(\linewidth - 6\tabcolsep) * \real{0.2500}}@{}}
\toprule\noalign{}
\begin{minipage}[b]{\linewidth}\raggedright
model
\end{minipage} & \begin{minipage}[b]{\linewidth}\raggedright
idio same-type pctile (med)
\end{minipage} & \begin{minipage}[b]{\linewidth}\raggedright
types \textgreater{} 90th pctile
\end{minipage} & \begin{minipage}[b]{\linewidth}\raggedright
blue-op AUC, clinical (med)
\end{minipage} \\
\midrule\noalign{}
\endhead
\bottomrule\noalign{}
\endlastfoot
phi-4 & 75th & 10.7\% & 0.706 \\
Qwen3-8B & 72nd & 2.0\% & 0.661 \\
Llama-3.1-8B & 93rd & 52.7\% & 0.571 \\
\end{longtable}
}

The scaffold itself is enormous and type-indifferent (random cross-text
anchor pairs overlap at median 1.00 in the tuned models, 0.36 in base
Llama). Type-transfer separates the models into two groups, inversely to
their fidelity to the analyst's markings: the base model (half of whose
shared types transfer above the 90th percentile of the scaffold
baseline) is the corpus's one type-anchored reader, and its fidelity is
the worst; the two instruction-tuned models transfer near baseline and
read best. We report the two-group form deliberately: an earlier corpus
composition ordered the tuned pair one way on the median (phi below
Qwen), the present one orders it the other way; the within-pair ordering
is not robust, and we do not claim it. The between-group gap is large
and survives every composition we have tested. Where within-register
transfer does appear in a tuned model, it sits on the register's
conversational machinery rather than its content lexicon: the types the
fourth transcript sends above the 90th percentile in phi-4 are the
hedges, discourse markers and speech-act verbs of spoken group
discussion (glaube, denke, irgendwie, gesagt), while the thematic
signifiers pairing the same transcripts (Liebe) stay at baseline. The
shared speech situation recurs and the reading of its frame recurs with
it; nothing has to travel with the word.

The two sides of the alignment carry the same structure, and the
fidelity axis has to be read with both planes subtracted. The scaffold
is subtracted here; the lexical baselines are subtracted in §5, where
the increment survives cleanly in phi-4, weakly in the base model, and
is not adjudicable in Qwen3-8B, and where the readers' gap attenuates
from 0.141 to 0.052 between the two the method decides. The two-group
form is accordingly a claim about tracking the annotation as a whole,
not about the singular surplus alone. The matched pair is exempt by
construction: base and twin read the same transcripts over the same
plane, so the twin's gain is a fact about the reader. Where the gain
lands, that exemption alone cannot say, since tuning could in principle
re-couple the aggregator to mark-predictive lexical structure; the
comparison below decides it.

\textbf{The matched pair.} The matched-lineage follow-up (Methods) is
\texttt{Llama-\allowbreak{}3.\allowbreak{}1-\allowbreak{}8B-\allowbreak{}Instruct}, the base's instruction-tuned twin, same
pipeline and constants, every corpus text, reported here and pooled
nowhere. It returns a split verdict, and the split is the result. On
transfer the twin is its base: idiosyncratic same-type transfer at the
93rd percentile, 55\% of types above the 90th against the base's 53\%,
the scaffold overlap equally lineage-stable (0.40 against the base's
0.36, where the tuned pair saturates at 1.00). On reading it is a tuned
model: blue-operative AUC rises on all four transcripts (median 0.650),
the §1 register concentration appears (red 0.33 and 3.54), the §3 red
hole opens, while the anchor geometry stays its base's: recruitment
order (blue-strong later in 1/9 runs) and both trace-column signatures
of §7 (its aggregator flat across the expert's column groups; its
differentiator unreorganized by the framing paragraph, 0.92 like the
base). The §5 controls certify where the conferred fidelity lands. The
twin's residualized AUC exceeds the base's on every transcript (0.678,
0.622, 0.588, 0.640 against the base's 0.661, 0.582, 0.531, 0.582;
pooled 0.627 against 0.585), and its pooled increment clears both
permutation nulls (p = 0.0001 block, 0.0003 rarity-matched) where the
base clears the block null only. The added fidelity is not a re-coupling
of the aggregator to mark-predictive lexical structure, a mechanism the
corpus shows to be available to a tuned reader, since Qwen3-8B's
aggregator is coupled to that axis at 0.15--0.22 of its rank variance
(§5). The twin's is not: 0.007--0.035, against its base's 0.002--0.028
and Qwen3-8B's 0.153--0.223, while the residualized alignment rises.
Since that quantity is measured without the annotation, it settles the
mechanism rather than restating the result: in the matched lineage the
conferred fidelity lands in the singular remainder. Instruction tuning,
in the one lineage where the comparison is clean, confers the
discourse-reading competencies (clinical fidelity, register sensitivity,
the exclusion structure) and leaves the geometry untouched:
type-essence, recruitment depth and column organization are laid down by
the lineage, tuned or not. The two-group form of the table above
therefore decomposes: type-transfer separates \emph{lineages} (phi and
Qwen against Llama, tuned or not) and fidelity separates tuned from base
\emph{within} a lineage. What Figure 7 shows as one inverted axis is two
facts that align across these three readers; the twin, carrying
base-level transfer and near-Qwen fidelity, is the point off the axis.
The type-essence, where the corpus has one, is inert: reading skill is
added without touching it.

\subsection{7. Execution traces (Tier 2,
structural)}\label{execution-traces-tier-2-structural}

\begin{figure}[H]
\centering
\pandocbounded{\includegraphics[keepaspectratio,alt={Figure 8 --- column dissociation}]{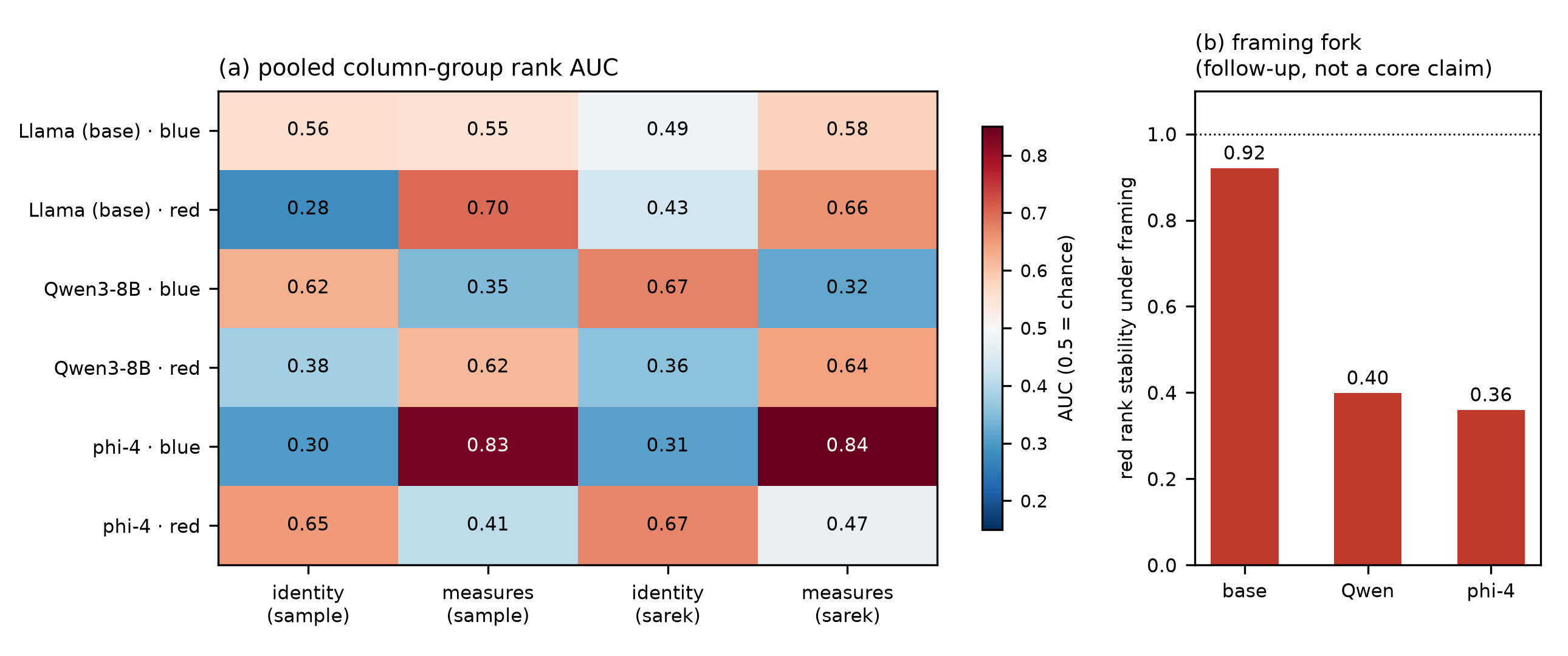}}
\end{figure}

\textbf{Figure 8 --- the channel\ensuremath{\times}model column
dissociation.} (a) Pooled column-group rank AUC (a group's tokens
against every other scanned token) on the two traces whose resource
fields carry real values: phi-4's aggregator on the measurement columns,
its differentiator on identity; Qwen3-8B inverted; the base model's
aggregator flat. (b) The framing fork (follow-up, not a core claim):
red-channel rank stability under a prepended explanatory paragraph ---
phi-4 and Qwen3-8B reorganize (0.36/0.40), the Llama lineage does not,
base and its instruct twin alike (0.92/0.92).

The second Tier-2 ground truth consists of four pipeline execution
traces (tabular records; 12 runs), for which a domain expert declared
the columns a domain reader consults (process name, status, and the
runtime/CPU/memory fields), the annotation fixed before any model output
was seen. This is a \emph{structural/data} ground truth,
methodologically distinct from the linguistic annotation of §5: it does
not test linguistic understanding in a direct sense, and it is scored
per column, never pooled with the clinical result.

The pooled marked-vs-unmarked AUC is uninformative (\ensuremath{\approx}0.5
in all cells): the interest set is roughly half the tokens and is
dominated by the long \texttt{name} field. Per column group, the
channels do concentrate on the expert's fields, but they split by
channel and by model. Grouping the declared columns into \emph{identity}
(name, status) and \emph{measurement} (the resource fields) and scoring
each group's rank AUC on the two traces whose resource fields carry
values (both replicated; \texttt{column\_\allowbreak{}transfer.\allowbreak{}py}):

{\def\LTcaptype{none} 
\begin{longtable}[]{@{}
  >{\raggedright\arraybackslash}p{(\linewidth - 8\tabcolsep) * \real{0.2000}}
  >{\raggedright\arraybackslash}p{(\linewidth - 8\tabcolsep) * \real{0.2000}}
  >{\raggedright\arraybackslash}p{(\linewidth - 8\tabcolsep) * \real{0.2000}}
  >{\raggedright\arraybackslash}p{(\linewidth - 8\tabcolsep) * \real{0.2000}}
  >{\raggedright\arraybackslash}p{(\linewidth - 8\tabcolsep) * \real{0.2000}}@{}}
\toprule\noalign{}
\begin{minipage}[b]{\linewidth}\raggedright
model
\end{minipage} & \begin{minipage}[b]{\linewidth}\raggedright
blue on identity
\end{minipage} & \begin{minipage}[b]{\linewidth}\raggedright
blue on measures
\end{minipage} & \begin{minipage}[b]{\linewidth}\raggedright
red on identity
\end{minipage} & \begin{minipage}[b]{\linewidth}\raggedright
red on measures
\end{minipage} \\
\midrule\noalign{}
\endhead
\bottomrule\noalign{}
\endlastfoot
phi-4 & 0.30--0.31 & 0.83--0.84 & 0.65--0.67 & 0.41--0.47 \\
Qwen3-8B & 0.63--0.67 & 0.32--0.35 & 0.36--0.38 & 0.62--0.64 \\
Llama-3.1-8B (base) & 0.49--0.56 & 0.55--0.58 & 0.28--0.44 &
0.66--0.70 \\
\end{longtable}
}

phi's aggregator lights the \emph{measurement} columns (0.83--0.84)
while its differentiator marks the \emph{identity} columns (0.65--0.67);
Qwen inverts the aggregator exactly (identity 0.63--0.67, measures
0.32--0.35). This is the §3 pin-material dissociation carried to an
external field list: Qwen's recurring-lexeme pins latch on the repeated
process-name strings, phi's structural aggregation on the numeric
fields. The base model shows \emph{no} such aggregator organization:
blue is flat across the two column groups (identity 0.49--0.56, measures
0.55--0.58, no preference), consistent with §6 (the most type-anchored
reader, which on this format-impoverished data organizes on the format
scaffold, headers and delimiters and identifier fragments, rather than
on content). The matched instruct twin (§6; never pooled) leaves this
flat: its blue holds no column preference either (identity 0.57--0.60,
measures 0.56--0.57), so the column organization phi and Qwen carry is a
lineage signature, not a product of instruction tuning.

\textbf{Attending to a column is not finding the heavy rows.}
Correlating the aggregator value against parsed resource magnitude per
row, runtime is the only field tracked with any consistency: blue
vs.~\texttt{realtime} magnitude is positive in three of the four tuned
cells (+0.50 \ldots{} +0.60) but near zero in the fourth (phi-4 on one
trace, +0.01), while memory and CPU magnitude are not reliably tracked
(signs mixed across traces and models). The channels locate the expert's
fields; beyond runtime they do not rank the expensive processes.

\emph{Follow-up direction:} A discourse-framing minimal pair, a light
explanatory paragraph prepended to a bare table, reorganizes the
\emph{differentiator} in phi-4 and Qwen3-8B (operative view; the
aggregator chain and the entire quiet view are invariant), moving red
off the arbitrary identifier columns onto the fields the frame names
(red channel stability 0.36/0.40). The Llama lineage does not
reorganize: base and its instruct twin both hold at 0.92, so this too is
lineage rather than a tuning effect (§6, matched pair). Because the
frame points by \emph{name} rather than by content, it does not by
itself improve alignment with the data where the named fields are empty.
We report this as an interesting follow-up, developed in separate work.

\subsection{8. Robustness checks}\label{robustness-checks}

\begin{itemize}
\tightlist
\item
  \textbf{Novelty baseline.} The channels are not difficulty in
  disguise. Against horizon entropy (the 27 non-trace runs), token-level
  correlations are moderate and genre-heterogeneous: on clinical
  transcripts, operative aggregator values correlate \emph{negatively}
  with entropy (\ensuremath{-}0.18 \ldots{} \ensuremath{-}0.64; pins
  sit where the horizon is most consolidated) while differentiator
  values correlate positively in two models; on the engineered technical
  texts both correlations are weak and sign-mixed (\ensuremath{-}0.29
  \ldots{} +0.52). The register results of §1 obtain on texts where the
  entropy correlation is near zero (\ensuremath{-}0.12, phi-4, WFA 1).
  Realized-token surprisal correlations are likewise weak and
  genre-split: on clinical transcripts the operative
  aggregator--surprisal correlation is near zero (median 0.02) and the
  differentiator--surprisal correlation mildly negative (median
  \ensuremath{-}0.16), and on the engineered texts both are weak and
  sign-mixed. The claim is token-level and does not extend to every
  grain: at region level entropy and the quiet aggregator co-peak on one
  marked disruption (§5), while on the engineered anomaly they separate
  in sign (§4).
\item
  \textbf{Attention saliency (post-hoc, descriptive).} The aggregator is
  not the model's own traffic. Attention is the saliency instrument an
  interpretability reader would reach for before any channel of ours, so
  it was scored on the clinical split of §5: per-token attention
  received and attention rollout (Abnar and Zuidema 2020), head-averaged
  over a fresh teacher-forced pass per transcript, each token's received
  mass divided by its number of attending queries so that the causal
  mask's favouring of early positions cannot pose as alignment
  (\texttt{attention\_\allowbreak{}saliency.\allowbreak{}py}; measures and corrections fixed
  before any value was computed, Methods). Raw, all nine pooled cells
  sit at chance, 0.47--0.51. Under the subtraction of §5 they rise
  rather than fall, to at most 0.56 within content words and 0.55--0.58
  residualized, because attention concentrates on the frequent material
  the analyst does not mark, so removing the lexical component uncovers
  a residue; residualizing helps a measure coupled negatively to the
  covariates and costs one coupled positively, and raw-against-
  residualized comparisons across two measures are not otherwise
  interpretable. That residue is level with the base model's
  residualized blue (0.576 against 0.585). It is also flat, varying by
  0.015 across the three models where blue's residualized values span
  0.518 to 0.637, and no attention measure has been put to the
  permutation nulls. Whether the readers' spread runs through it is
  tested directly: blue residualized on all three attention measures is
  unchanged in every model (base 0.568, Qwen3-8B 0.664, phi-4 0.712,
  against raw 0.562, 0.657 and 0.703), and adding attention to the
  lexical covariates never lowers the calibrated increment by more than
  0.013 (\texttt{outputs/\allowbreak{}comparison/\allowbreak{}attention\_\allowbreak{}aucs.\allowbreak{}csv}). The two also
  part under return: across the 33 term \ensuremath{\times} model cases of
  §5 the rollout percentile falls from first to last occurrence in 32,
  blue in 21. A token's consolidation and the volume with which it is
  read from are different quantities on this split (Discussion,
  ``Structural value'').
\item
  \textbf{Pins are not sinks (post-hoc, descriptive).} The two measures
  are easiest to conflate on the delimiter crust of §3, so they were
  scored against each other there. Delimiters do draw more attention
  than other tokens, in all twelve cells (median received percentile
  0.61--0.65 by model against 0.46--0.47 for non-delimiters). They are
  not the same delimiters: splitting the delimiter class by whether the
  aggregator places the token in its top decile, the pinned positions
  sit \emph{below} the unpinned ones in received attention in both tuned
  models (phi-4 0.50 against 0.70, Qwen3-8B 0.36 against 0.67) and level
  in the base model (0.58 against 0.60), the ordering holding in all
  seven testable tuned cells and in two of the base model's four
  (Qwen3-8B's fourth transcript pins no delimiter, so eleven cells are
  testable). The two top deciles overlap at 0.04--0.11 where chance is
  0.10, and the token-level correlation is null to negative (base +0.03,
  phi-4 \ensuremath{-}0.15, Qwen3-8B \ensuremath{-}0.33). The
  paragraph-final pins sit at the 0.98 aggregator percentile by
  construction and the 0.64--0.71 attention percentile in phi-4.
  Received attention is divided by the number of attending queries, a
  correction that favours late positions rather than penalizing them, so
  the pins' middling attention is not made by the correction. The one
  position where the sink literature does describe the aggregator is the
  first token (below), which no headline result uses. Scope: the
  clinical split only, on which the attention pass was taken; the
  capiton test of §3 runs on the discovery text and has no attention
  counterpart.
\item
  \textbf{Delimiter erasure (post-hoc, descriptive).} The aggregator is
  not punctuation in disguise. Stripping every delimiter token from the
  scoring pool raises the clinical blue-operative AUC in all four
  transcripts of both delimiter-heavy models: phi-4 to 0.72--0.77 (from
  0.68--0.72), the base model to 0.52--0.72, its single below-chance
  cell (H3, 0.470) moving above chance to 0.522. It lowers Qwen3-8B's by
  \ensuremath{\approx} 0.04 (to 0.58--0.66): its delimiters sit low in the
  ranking, where they served as easy negatives. The delimiter crust of
  §3 masks the alignment rather than producing it. The registered 34/36
  count stands as scored; this rescoring is post-hoc.
\item
  \textbf{First-position elevation.} Consistent with the attention-sink
  literature, the first valid token shows mildly elevated aggregator
  values (median 73rd percentile across the 27 non-trace runs;
  top-decile in 30\%); the differentiator is not elevated (18th). No
  headline result depends on first-position anchors: excluding the first
  k opening anchors from the readout (k = 1, 2, 4, 8) leaves the
  boundary directions (6/6) intact and shifts the clinical
  blue-operative AUCs by at most 0.0031 across all 12 cells and all k,
  with the pin-identity shares moving by at most 0.04 (recomputed by
  \texttt{paper\_\allowbreak{}stats.\allowbreak{}py}). The paragraph-final capiton pins sit deep
  in the text and are untouched by any such exclusion.
\item
  \textbf{Determinism and self-healing.} All numbers regenerate from
  cached passes with one command; scan artifacts carry definition tags
  and recompute when definitions change; prefix state identity (1.0)
  bounds float noise.
\item
  \textbf{Boundary placement} is exact to the token under both tokenizer
  families (no leading-space off-by-one).
\end{itemize}

\section{Discussion}\label{discussion}

\subsection{Structural value}\label{structural-value}

When the same wording is read after two different pasts (the shared
suffix of the edited and base versions), the tokens' \emph{states}
barely move: median final-layer cosine between versions is 0.987--0.993
near the edit and 0.996--0.998 at distance, against a measured identity
floor of 1.0 on the shared prefix. Their \emph{structural values} are
also largely preserved (the channels' Spearman rank correlation over the
same aligned tokens is 0.73--0.97), but the two stabilities are not of
the same order: where the states sit within one percent of their
identity ceiling, channel rank stability drops by up to a quarter, and
the drop concentrates in the differentiator channel of the
instruction-tuned models. A distant edit that leaves a token's substance
nearly untouched measurably renegotiates its position in the text's
organization. Same substance, slightly different system of differences,
disproportionately different value.

This is, we take it, evidence against what might be called the
essentialist reading of hidden states: the assumption that a token's
representational content, its metaphoricity, its importance, its
structural function, inheres in its vector, waiting to be read off by a
sufficiently good probe. Were that so, value should be exactly as stable
as the vector it inheres in; instead it is systematically less so, and
the shortfall tracks the relational channels. What our measurements
exhibit is closer to Saussure's distinction between \emph{signification}
and \emph{valeur}: a signifier's value is constituted by its differences
from the co-present others, not by intrinsic content. The anchor
channels implement this constitutively, not accidentally: a basin is a
set of \emph{other} tokens, a pull gap is a difference between
populations, an aggregator's rank is a property of the whole
configuration. There is no one-token version of ``being an anchor.''

Note that within the embedding the claim is stronger still than in
Saussure's linguistics. The signifier, for Saussure, still faced a
signified; a token's vector faces only other vectors. Its intrinsic
content is, exhaustively, a position, and a position is nothing but the
relations it enters, so reference does not anchor value even in
principle: in the residual stream there are only signifiers. An
essentialism about hidden states accordingly needs more than the
stability of the vector; it needs an alignment claim between the
embedding space and the world. Such a claim was ventured for a time in
interpretability research (the ``Platonic representation hypothesis,''
on which representational convergence across models is read as
convergence on reality; Huh et al.~2024), but what converges there is
representations with representations, models trained on overlapping
corpora agreeing with one another: convergence among readings, not
contact with a referent.

\textbf{Value against information.} Saussure supplies the synchronic
half of this claim; its diachronic half is Lacan's, and the clinical
tier gives it numbers. In the theory this paper operationalizes, the
weight of a signifier comes from its position in the chain and from its
return, the insistence of the signifying chain (Lacan 2006), not from
its informativeness. Shannon-side reasoning orders the same tokens the
other way: weight as improbability, and improbability is what return, by
construction, destroys. The two accounts therefore part ways on a
measurable sign wherever a marked signifier repeats, and Results §5
returns the Lacanian sign. Over the split as a whole the predictive
measures anti-align, both below chance; at the type level
informativeness survives only as rarity, the generic share every reader
gets for free; and at the level of the occurrence, where the theory
locates the weight, the dissociation is clean, a marked word's surprisal
percentile collapsing across its repetitions while its aggregator
percentile holds. The shape is systematic, not curated: over every
marked term that repeats, the median first-to-last change is
\ensuremath{-}0.16 in surprisal percentile against \ensuremath{-}0.02
in blue; return drains the one and leaves the other standing. The
analyst marks the signifier because it returns; an information reader
must lose it for the same reason; the position-in-configuration measure
stays with the analyst. We registered no such prediction. The theory
stated it seventy years before the instrument existed, and the
instrument returns its shape in the registered channel. The dissociation
has a second edge that sharpens it. One might grant that return drains
predictability and still suspect the aggregator of tracking nothing more
than how heavily a token is attended to, its salience in the model's own
traffic. It does not. Raw, the standard saliency measures sit at chance
on the clinical split in all three models; put under the same lexical
controls the channel is held to it rises, to 0.55--0.58, since attention
drains toward the frequent material the analyst leaves unmarked and
removing that component uncovers a residue. So attention does carry a
share of the marks, and the dissociation is not that it reads nothing.
It is that the residue is flat across the three readers where the
aggregator's is not, and that the aggregator's own alignment survives
residualization on attention undiminished (Results §8). What it cannot
do at all is hold under return: scored across the same repetitions it
drains exactly as surprisal does (median first-to-last change
\ensuremath{-}0.18 in rollout percentile, against \ensuremath{-}0.02
in blue; Results §5). A token read from ever less as it insists is still
consolidated by the aggregator; the volume of traffic to a signifier and
its weight in the configuration come apart on the same marks where
predictability does. What the channel holds is not the information the
token carries, nor the attention it draws, but its standing in the
chain, and only the last of the three stays with the analyst.

The engineered anomaly supplies the converse corner, and it was built
before any of this was measured. Return drains informativeness and the
aggregator holds; the absurd passage maximizes informativeness and the
aggregator leaves. Its tokens are the most improbable in the text
(predictive entropy 0.63--0.67, word surprisal 0.63--0.65, above chance
in all three models) and the aggregator is at 0.17--0.32, below chance
in all three, at a position the junction control clears (Results §4). A
quantity that measured information would have to rise in both places;
one that measures standing in the chain falls exactly where a passage,
however startling, is woven into nothing. The static lexical predictors
are indifferent there (0.49--0.54), which is the same point from the
other side: what the passage lacks is not unusual words but a place.
There are two qualifications to mark the limits of this approach.

\textbf{First, instrumental vs.~representational relationality.} Part of
the observed reordering is our own ruler: the channels normalize by
within-text quantiles and whole-pool medians, so some instability is
measurement-relative. The objection has an essentialist form: ``the
states stayed at 0.99; only your instrument is relational.'' But the
individuality is in the substrate, prior to any ruler. Each
residual-stream state is a function of its entire causal prefix, so the
geometry is text-individual by construction of the architecture, not by
our scoring, and the minimal pair shows it cleanly: causal masking holds
the shared prefix \emph{exactly} fixed (cosine 1.0 to float precision, a
consequence of the mask, not a measured near-miss), while the suffix,
recomputed against a context that now contains the insertion, moves in
the raw states themselves, before a single channel is taken. The relata
are not merely different in count; the architecture individuates them.
What normalization adds on top could be partitioned by rescoring against
a matched pool (beyond the present scope), but the raw-state divergence
already settles the objection: the state vector underdetermines
structural value.

\textbf{Second, the claim is not that structural value is unstable.}
Channel rank stability of 0.73--0.97 means a token's role largely
survives a distant edit; both halves of the result, near-ceiling state
stability \emph{and} high-but-lower value stability, belong in the same
sentence whenever this finding is cited. The licensed claim is
comparative, not absolute: value is \emph{more context-sensitive than
the representation it is read from}, by roughly an order of magnitude in
the corresponding similarity gaps, and most sensitive exactly where the
measure is most relational. What is absolute in the residual stream is
the token's state given its causal past. What is not is what that state
\emph{counts as} within a text's organization, and the difference is
measurable.

Note, finally, what did not dissolve. Values are relative within a text
(the suffix result) and relative across models, the same organizing
function is installed in categorially different material, delimiter
tokens in one model, recurring lexemes in another. Yet the
\emph{partitions} replicate: the engineered-boundary direction in 6/6
text\ensuremath{\times}model cells, the anomalous passage's signature,
aggregation void, elevated operative-view differentiation, in 3/3
models, and content- rather than position-specific under the junction
control. Anti-essentialism about token meaning is usually taxed with
relativism: if value is nowhere intrinsic, anything goes. The corpus
pays that tax in advance. Differential systems, each running on its own
material, converge on shared structure; the value of a token is nowhere
absolute, and the structure of differences is nonetheless robust enough
to be found by three architecturally unrelated readers, and to agree
with where a domain expert says the information is.

Nor is value smuggled in by the \emph{type}. If the relationality
argument holds, a word recurring in two texts should carry nothing
privileged from one to the other beyond what the trained scaffold gives
any token; if token meaning were a form in the weights, same-type
transfer is exactly where it would surface. The type-essence test
(Methods, battery test 7) measures this directly, and the answer is
graded: with the substrate removed, same-type cross-text subspace
overlap sits at the 72nd--75th percentile of a random-anchor baseline in
the instruction-tuned models and at the 93rd in the base model, with the
share of types exceeding the 90th percentile at 2--11\% in the tuned
models against 53\% in the base. In the tuned models the same signifier,
read in two discourses, is for the most part simply two anchors. What
persists across texts everywhere is enormous but type-indifferent: the
scaffold, a trained plane available to every token alike.
Methodologically this cashes out as a warning: any cross-text statistic
pooled by lexical type would average away precisely the text-individual
structure under study.

The dichotomy is the finding: the two-group claim. Across the corpus's
three readers, the more essentialist the geometry, in the operational
sense that a larger share of its anchor structure is carried by the
lexical type across contexts, the worse it reads the singular discourse,
as a contrast between groups, not a ranking within them. The mundane
baselines (Results §5) condition this contrast without dissolving it:
with the generic plane subtracted the gradient attenuates, clearest in
phi-4, weaker in the base model, not adjudicable in Qwen3-8B, the gap
between the two the method decides falling from 0.141 to 0.052. The
matched pair is exempt from the caveat by construction, base and twin
reading over the same lexical floor, and the residual comparison decides
the mechanism: the twin's residualized increment exceeds the base's on
all four transcripts and clears both permutation nulls, while its
channel's coupling to the lexical axis stays in its base's weakly
coupled regime, so the fidelity tuning confers lands in the singular
remainder, not in a lexical re-coupling (Results §6). It also decomposes
the contrast into two facts that the corpus had fused. Instruction
tuning does not de-essentialize the anchor geometry: the instruct twin
transfers exactly as its base does, at the 93rd percentile, 55\% of
types above the 90th against the base's 53\%. What tuning moves is the
reading: the register concentration, the channel exclusion, above all
the fidelity to the analyst's marks, up on all four transcripts to
within reach of the tuned pair. Type-anchoring is constitution of the
lineage, laid down before any instruction arrives; the fidelity is
conferred over it without touching it. This sharpens the relational
point rather than weakening it. The clinical signifiers the expert marks
are constituted by their position in \emph{this} patient's speech, not
by their dictionary identity, and the one geometry in the corpus that
demonstrably carries a type-essence carries it inertly: fidelity to the
singular discourse is gained, in the twin, with no change in what the
type transports. Whatever the essence stores, it does no reading work;
the reading that tracks the expert lives entirely in the layer tuning
rewrites. An essentialist account now has two findings to explain away:
the corpus correlation, in which fidelity to type-forms accompanies
insensitivity to discourse, and the within-lineage dissociation, in
which the sensitivity is added while the type-forms stand unmoved.

\subsection{The trace of rupture}\label{the-trace-of-rupture}

That the marked disruptions leave channel traces at all (and only as
extremes against a sliding-window base rate, not as mere deviations) is
the finding of Results §5. The collapsed-syntax passage of transcript 3
is the most quiet-blue-elevated window of its transcript in all three
models, transcript 2's principal marked passage the most withdrawn: the
analyst's marks sit at the far ends of their transcripts' distributions,
in specific and replicated cells. Where the signature does not replicate
(transcript 1, and the second span of transcript 2), the alignment is
one model's; the fourth transcript's single marked passage is treated
below. The annotation itself carries no types: the analyst marked only
\emph{that} the discourse breaks. Everything beyond this is read off the
geometry, not off the marks, and is correspondingly provisional.

The traces are not uniform, and their differences suggest kinds: stated
here as candidate signatures (strong-anchor density inside the passage
\ensuremath{\div} outside), not as an established taxonomy. The
engineered anomaly, fluent but false (coconut-water furnaces, dolphins
measuring purity), shows elevated operative \emph{differentiation} (a
red rim, 2.4\ensuremath{\times} in phi-4) around an aggregation void, with
silence in the quiet web. A grammatically broken, self-interrupting
passage (transcript 3: collapsed syntax, aborted clauses) elevates the
quiet web in all three models (2.2--2.9\ensuremath{\times}, the transcript's
most elevated window of its size in each model), with an operative red
rim in two of three. The two fluent passages the analyst nonetheless
marked withdraw from quiet-view aggregation in five of six cells
(0.42--0.61\ensuremath{\times}; the sixth, phi-4 on the idealization
passage, sits at 0.88), transcript 2's span being its transcript's
single most withdrawn window in all three models, with weak and
model-dependent operative involvement (red 0.64--1.27\ensuremath{\times});
that the disruption in these passages lies in the content rather than
the form is our reading, not the annotation's. The fourth transcript's
single marked passage is the exception, and what it withholds at one
grain it shows at another. The span the analyst marked there is long
enough that it deviates in no cell against its surroundings, and too
long for its transcript to offer enough same-size windows to rank it in
a distribution at all. Segmented at the speaker-turn boundaries, a
segmentation that is ours, not the annotation's, recorded as superseded
in the annotation file, the sub-spans trace in opposite directions that
cancel in the aggregate: the interjection elevates the quiet web in two
of three models (1.56\ensuremath{\times}, 2.10\ensuremath{\times}), while the
same speaker resuming after it withdraws with the most uniform quiet
trace in the corpus (0.706--0.708\ensuremath{\times}, 3/3 models). The
instrument misses the disruption at the grain of the analyst's mark and
differentiates it at the grain of the exchange. Whether the channels
would find those boundaries unaided is a designed test (boundary-blind
changepoint detection on the stored trajectories) that lies beyond the
scope of this study.

The scan that supplies the base rate also returns the converse: unmarked
windows whose signature replicates in all three models. The strongest
lands in transcript 3 and carries that transcript's own marked
signature, quiet-blue elevation, 3/3, on a passage of visibly broken
speech: long pauses, self-interruption, and syntactic breakdown (``dass
(\ldots.) dass ich fühle mich nicht so''). Two more of the same kind sit
on an aborted word cut off by a speaker change (transcript 1) and on a
halting, unfinished search for words (transcript 2, both views
withdrawing). We read these as candidate ruptures, not as false
positives: the annotation marks one clinical reading of a longer text,
and an extract has to stand on its own, the region-level counterpart of
the unmarked lattice below. What the candidates share with the marks,
audible disturbance of the speech surface, is again our reading, and the
designed test is the same larger, pre-registered corpus the typology
already waits for.

A mechanistically economical hypothesis organizes these differences:
operative differentiation tracks surface-processing friction,
indifferent to clinical meaning: it rises for nonsense and for broken
grammar, and tends toward flat where the surface is smooth however
loaded the content. The red channel would then know friction, not
significance, and a fluent defense would register, if anywhere, as a
quiet-web deficit, the one place a purely form-driven measure would not
look. We flag where this is already strained: phi-4 puts rim-level red
(1.27\ensuremath{\times}) on one fluent marked passage, so the clean form of
the claim, fluent passages lack the rim, fails in at least one cell.
Whether that cell is noise or the beginning of a fourth kind is exactly
what the present corpus cannot decide: seven disruptions,
single-annotator. A substantially larger transcript with further marked
ruptures already exists in the project, and the typology question, with
the friction hypothesis as its testable core, belongs to a second,
clinically directed study built on it.

\subsection{The isotropy tension}\label{the-isotropy-tension}

Timkey and van Schijndel (2021) showed that removing the rogue
high-variance dimensions aligns model similarity spaces \emph{better}
with human similarity judgments, and the isotropy-enforcement literature
generalizes the moral: the substrate obscures what humans see. Our
Tier-2 result points the other way: it is the \emph{raw, operative} view
whose aggregator channel aligns with the analyst's markings, while the
whitened view anti-aligns. There is no contradiction, but the resolution
is informative. Similarity judgments ask a human to compare words as
\emph{types}, out of discourse; structural markings ask an expert to
locate organizing points \emph{within} one discourse. Whitening helps
the first task because type identity lives in the quiet dimensions (our
quiet view is where a text's recurrent lexical web registers); it hurts
the second because discourse-level organization is computed on the
substrate, the geometry the model's own attention arithmetic runs on.
The two literatures are measuring different human competences against
different strata of the same space, and the two-view design makes the
division visible instead of averaging over it.

\subsection{Ground truth without
agreement}\label{ground-truth-without-agreement}

A standard paradigm of computational linguistics would ask, at this
point, for inter-annotator agreement: a second analyst, a kappa, a
``ground truth'' certified by convergence. For the clinical tier we
reject the demand, and the rejection is methodological. Agreement
statistics presuppose that the annotation target exists as an
observer-independent semantic layer over which independent judges
classify. The organization of a singular subject's speech, its quilting
points and its ruptures, is no such layer: it is relational through and
through, the economy of one discourse, and reading it is interpretation,
not classification. To demand that two analysts produce identical
markings is to mistake the one for the other; and perfect agreement,
were it achieved, would count against rather than for the annotation,
since point-for-point convergence is what surface lexical cues produce,
shared signification, the Imaginary register, not the structural economy
of this patient's speech. The mundane baselines of Results §5 give this
claim its quantitative body rather than contradicting it: the share of
the annotation on which any reader, human or statistical, converges is
now measured, a frequency list reproduces the marked/unmarked split at
0.88, and it is precisely the share on which a kappa would be computed
and rewarded. What individuates the analyst's reading is the remainder:
not \emph{that} rare content words organize a discourse, but
\emph{which} of them, and which occurrences insist. The perspectivist
turn in NLP has reached a neighboring insight from the opposite
direction (Aroyo and Welty 2015; Uma et al.~2021; Plank 2022): label
variation is signal, not noise, wherever the task is interpretive.
Psychoanalytic annotation is the limit case: variation there is not
merely tolerable, it is constitutive of the object. In line with
perspectivist NLP, we treat the analyst's markings as a situated expert
probe.

Three constraints keep the rejection from becoming a license. First,
validation in this paper never rested on inter-subjective convergence:
the marks predate the instrument, the alignment claim is directional and
replicated across three architecturally unrelated readers, and every
clause could have failed. Second, the two Tier-2 grounds part ways
exactly here: declaring the columns of interest in an execution trace
\emph{is} a classification, agreement is meaningful there, and the
separation of the tiers maintained throughout the paper is now visible
as an epistemological distinction, not a bookkeeping one. Third, and
decisively, the rejection is not imported from the theory but returned
by the data, in matching form on both sides of the alignment. On the
annotation side, the mundane baselines measure the shared plane
directly: what a frequency list recovers from the same split is
invariant across the three readers to within 0.02. That invariance does
not on its own consign the reader differences to the remainder, since a
channel can differ from another precisely in how far it rides the shared
plane, and one of ours does. The controls are what separate the two, and
they leave a reduced reader difference standing: the residual increment
and the signed dissociations of Results §5, graded clean in phi-4, in
the instruct twin, whose residual increments clear both permutation
nulls, and more weakly in the base model, undecidable in Qwen3-8B, whose
aggregator is the one measurably coupled to the generic stratum (Results
§5--6). On the model side, the type-essence tests show that the models
tracking the clinical annotation best are precisely those in which
structural value does not transfer by lexical type across texts: what
aligns with the analyst beyond the shared plane is the singular reader,
not the type-classifier. The geometry itself is constitutively
relational, and the annotation that tracks it must be read the same way.
A second analyst therefore remains valuable, but as a second singular
reading for the instrument to track, reader by reader, not as a
convergence statistic.

The point is not confined to human readers, which would make it easy.
Our three models are readers too, and each reads differently; an
agreement statistic across them would be perfectly computable, and it
would remove exactly what it is computed over, the individual reading,
the object of this paper. That is why no result here is an average over
models: what replicates is the form of the answer in each singular
reading, never a pooled magnitude. And in the models, the distinction
that in human readers can only be inferred (shared language on one side,
individual metaphoricity on the other, the one shifting the grounds of
the other) becomes directly measurable. The shared plane has a name and
a size: the trained scaffold, the type-indifferent overlap that any two
anchors share, and the transfer test subtracts it. What remains above
that baseline is the individual reading, and it is what tracks the
analyst. Agreement methods, human or machine, compute on the shared
plane; the object of this paper lives in the remainder.

\subsection{Processing metaphoricity}\label{processing-metaphoricity}

The differentiator channel was built on the ever-recruited basin, and
the interpretive weight of that choice deserves its own statement. A
token belongs to an anchor's basin if its state crossed the anchor's
threshold at \emph{any} depth, whether or not the alignment survives to
the final layer. Everything the red channel measures inherits this
transience: it marks positions at which depth-transient alignment,
transport in flight, occurred somewhere in the model's reading, not
positions whose resting representation carries a trace of it.

Three observations: First, recruitment is genuinely a depth event:
ordinary anchors recruit at a median layer of 17/40 (\ensuremath{\approx}
42\% of depth), and for the strongest aggregators basin membership
saturates mid-stack while final-layer cosines settle well below
threshold. The object the red channel measures has, at the output,
largely ceased to exist. Second, the engineered anomaly is distinguished
by what its tokens \emph{passed through}, not by what they became: its
signature is a rim of elevated operative differentiation (up to
2.4\ensuremath{\times} strong-density) around an aggregation void. During
reading, the passage's tokens moved distinctively into idiosyncratic
subspaces; in the final configuration, nothing consolidates there.
Third, the suffix result: under a distant edit, the largest
channel-stability drops concentrate in the differentiator of the tuned
models (0.73, 0.76), as they should if the channel marks an event of the
reading, renegotiated whenever the past changes, rather than content
deposited in the token.

Together the three fix the predicate. \emph{Processing} metaphoricity is
transport as an event of the reading, not content deposited in a sign.
This is also why the channel cannot, and should not, support the claims
usually made under ``metaphor detection'' (e.g.~Aghazadeh et al.~2022;
Fuoli et al.~2025; Ichien et al.~2024): it asserts nothing about a
word's stored metaphoric content, its behavior under paraphrase, or its
type identity across contexts; the type-transfer result (above) says
such content largely is not there to be detected, in the models that
read best.

What would license final-state claims is precisely what our
correlational protocol excludes: interventions, patching an anchor's
core subspace would test whether a transported representation is
\emph{used}, a probe trained on resting states whether it is
\emph{legible}. The claim ``this text's reading passed through
metaphoric transport here'' and the claim ``the model can subsequently
use what was transported'' are different claims; we make only the first.
But consider the second one a sensible continuation of research under
our conceptual regime.

\subsection{The pin function}\label{the-pin-function}

The top-decile aggregator anchors, the pins, earn a functional
description that is stable across models whose pin \emph{material}
differs categorially: delimiters in phi-4 and the base model
(paragraph-final ones where the format provides paragraphs), the text's
recurring lexemes and content words in Qwen3-8B, which places no
delimiter at the top on any text. The counts describe the crust of a
ranking, not its body (the resistance to statistical erasure we demand
of ground truth applies to our own aggregates), and it is the content
stratum beneath phi-4's delimiter crust that meets the analyst's marks;
removing the crust strengthens the alignment (Results §8). The function,
wherever it is installed: a small set of token positions whose core
subspaces support a broad, high-rank configuration of the whole text
(capacity\_share), whose density tracks the technical register across
engineered boundaries in 6/6 cells including the reversed-order control,
whose basins saturate late (median recruitment layer 29/40,
\ensuremath{\approx} 72\% of depth, against 17/40 for ordinary anchors), and
whose values align with where a psychoanalyst independently marks the
organizing signifiers of a discourse (34/36 directional Tier-2 cells,
with the alignment's residue beyond lexical rarity calibrated in Results
§5). Where the material permits paragraph-final pins, they bind their
own preceding paragraph above the rest of the text (phi-4 on the
discovery text: operative \ensuremath{\Delta}cos +0.022, 14/18 pins, p =
0.031; quiet \ensuremath{\Delta}cos +0.066, 15/18, p = 0.0075).

The mechanistic literature explains why such positions are
\emph{available}: attention-sink work documents delimiters as
preferential aggregation sites, and SepLLM demonstrates that separator
tokens can carry their segment's content well enough to substitute for
it. Availability is not identity, and the two part company exactly where
they are easiest to conflate. Delimiters do draw more attention than
other tokens, in all twelve clinical cells (median received percentile
0.61--0.65 by model against 0.46--0.47), but they are not the same
delimiters: within the delimiter class the pinned positions sit
\emph{below} the unpinned ones in received attention in both tuned
models, in all seven testable cells (phi-4 0.50 against 0.70, Qwen3-8B
0.36 against 0.67), the base model is level at 0.58 against 0.60, and
the two top deciles overlap at chance (Results §8). Where phi-4 quilts,
it quilts at the delimiters attention passes over. So the pin is not the
sink under another name, and our contribution is not a re-description of
that mechanism but a different quantity: not that delimiters aggregate,
but that \emph{which} positions aggregate, with what capacity, maps an
individual text's organization, and that the map agrees with expert
structure.

It is this function that we propose reading under the master-signifier /
point-de-capiton vocabulary, and the analogy is functional, with stated
falsification conditions, not decorative. A quilting point, in the
theory, (i) retroactively fixes the meaning of the chain it terminates,
(ii) does so as a piece of signifying material that is itself
comparatively empty, its power positional rather than semantic, and
(iii) is master of \emph{this} discourse, not of the language. Each
clause is operationalized and each could have failed: (i) the
paragraph-binding test asks whether the pin binds precisely its own
preceding segment rather than the text at large, a pin that bound
indiscriminately, or bound its future, would break the analogy (the
directional split gives the unambiguous reading only for the past side);
(ii) the material clause is carried by the cross-model dissociation, in
phi-4 the function installs itself in
\texttt{.\allowbreak{}\textbackslash{}n\textbackslash{}n}, a token with no lexical
content at all, and the function replicates in Qwen in different
material, so what does the binding is position in the structure, not
content; (iii) the type-transfer result shows the pin function does not
travel with the token type across texts in the tuned models, mastery is
not a property the word carries between discourses. One limit of the
analogy is already visible in the data: Qwen produces no paragraph-final
pins, so clause (i) is untestable there. The claim is therefore
per-model and functional: where a model quilts with delimiters, the
delimiter-pins bind their segments. It is not a claim that every model
quilts at paragraph ends.

\subsection{The unmarked lattice}\label{the-unmarked-lattice}

The analyst's marks never exhaust the strong-aggregation set, and the
remainder is not noise. An exploratory pass
(\texttt{unmarked\_\allowbreak{}strong.\allowbreak{}py}; operative view, battery threshold, no
claim scored) removes from each transcript's blue-strong set the tokens
on analyst terms and the paragraph-final pins; three strata remain.
First, discourse machinery: in phi-4 and the base model roughly 60--75\%
of strong aggregation sits on bare delimiters (0--3 anchors in Qwen3-8B,
whose lattice is lexical), and where this stratum is verbal it is
connective: the only spans strong in all three models of transcript 1
are two occurrences of \emph{dass}, the hinge on which an assertion
becomes an embedded clause. Second, the other speakers' mirroring
discourse, consistent with the polyphony attenuation of Results §5.
Third, a small cross-model consensus of clinically legible but unmarked
signifiers (18 spans over the four transcripts), in transcript 4,
\emph{meine Mutter} in the abortion sentence, \emph{tiefe Enttäuschung},
and \emph{stark gedeutet}, the patient's own name for their interpretive
activity. The caveats are those of any exploratory residue, subword
granularity, a top-decile cutoff, one-occurrence term matching, and we
score no claim on it: the annotation marks what mattered to a clinical
reading of a longer text, and the residue is not a miss list.

The first stratum invites an interpretation. No human reader would mark
the lattice: it is the implicit structure of the text, exactly what must
remain transparent for reading-as-meaning to proceed. The instrument has
no transparency to preserve: it weights whatever does structural work,
and so renders the implicit as the most visible layer. Since the blue
channel is capacity share, ``the lattice carries the information'' is
close to a literal reading of the measurement: aggregation banks a
stretch of discourse at the articulation point that closes it. This
inverts the form--content schema classical linguistic theory assumes,
form as container, content as payload: what that schema calls form is
here an aggregation function whose token material is contingent, and
form, so measured, is impoverished discourse, discourse compressed to
its articulation, not a vessel distinct from what it carries. The claim
is functional because the material version fails in both directions,
Qwen3-8B showing delimiter material is not necessary and the base model
that it is not sufficient (Results §7).

The execution traces (Results §7) provide the degenerate control: where
discourse thins to a table the residue does not vanish but becomes
confused discourse, the cross-model consensus falls entirely on
timestamps and process-name strings (16 spans over the four traces), the
indexicals a reader would need to tell the table as a story, while the
quantities the rows exist to record are barely tracked. Reading
machinery built for discourse, applied to format, recovers the skeleton
of a discourse that is not there. The inversion holds from both sides:
in the transcripts the discourse banks itself in its lattice; in the
tables the lattice is nearly all the discourse there is.

We do not elevate this to a finding, single annotator, extract-level
marks, exploratory thresholds. It is stated because it is interpretable
now, and because it fixes the design of the study it calls for:
pre-registered residue predictions on held-out annotated texts, across
formats, at a scale of corpus and more importantly compute beyond the
present study.

\subsection{Two views, two strata}\label{two-views-two-strata}

The two-view design is one of the methodological spines of the paper:
the views disagree, and the disagreement is patterned. The boundary
alignment does not survive z-scoring: it \emph{inverts}, in both models
with complete quiet scans and in both engineered boundary directions
(operative blue tracks the technical register at 3.0\ensuremath{\times};
quiet blue tracks the soft register, 0.08--0.10 and 1.79--2.41 under the
reversed design). Per-anchor channel values correlate weakly or
negatively across views. And the rupture typology is legible only with
both views open at once: the anomaly is an operative rim over quiet
silence, the broken-syntax passage elevates both, and the fluent marked
passages register chiefly as a quiet-view withdrawal.

Our positive interpretation assigns the strata. The operative view is
the geometry the model's attention arithmetic actually computes on,
dominated by the anisotropic substrate; it is where discourse-level
mastery registers, the pins' capacity, the technical register's
consolidation, the alignment with the analyst's markings (the Tier-2
channel that aligns above chance is blue-\emph{operative}, while
quiet-blue anti-aligns in 12/12 cells). The quiet view exposes the
low-variance stratum in which the text's recurrent lexical web lives; it
is where local, segment-scale binding is most legible (the capiton test
is stronger in the quiet view: \ensuremath{\Delta}cos +0.066 vs +0.022) and
where a text's failure to weave a passage in shows as withdrawal. The
slogan version: organization-as-mastery on the substrate, texture and
binding in the quiet web. A single-view instrument would have reported
one stratum and been wrong about the other: the whitened-view-only
analyst would call the technical register unstructured; the raw-view-
only analyst would miss most of what the fluent marked passages do.

Consequently, every finding in this paper is claimed in the view that
carries it, and nothing is averaged across views: the views are not
repeated measures of one quantity but measurements of different strata,
and pooling them would manufacture noise out of signal. Conversely, a
finding visible only in the quiet view is a finding about a
\emph{rescaled} geometry: per-layer z-scoring is our instrument, not the
model's computation, and quiet-view claims are correspondingly claims
about what structure exists in the low-variance dimensions, not about
what the model's attention consumed. Where a result is carried by the
operative view, as the expert-alignment results are, this caution is not
needed, which is one reason we consider the controlled Tier-2 outcome,
the increment and dissociations that survive the §5 baselines, the
paper's most robust external anchor.

\subsection{The global workspace}\label{the-global-workspace}

Gurnee et al.~(2026) recently identified, with a corpus-averaged causal
instrument (the Jacobian lens), a sparse set of verbalizable directions,
the J-space, that behaves like a global workspace in Claude-family
models: roughly 10--25 concept directions active at a time, under 10\%
of activation variance yet spanning most of the residual stream,
emerging with ignition-like sharpness around a third of the way through
the stack, persisting across nearby positions, and giving way in roughly
the last tenth of layers to output-tied ``motor'' representations.

The two instruments share nothing methodologically (theirs causal,
corpus-averaged, direction-space, vocabulary-tied; ours correlational,
per-text, subspace-based, token-anchored), which is what makes the
convergences structural: both find the content-bearing structure off the
dominant-variance axes, and both find the same depth articulation, which
the channel-exclusion result (Results §3) lets us state channel by
channel.

If either of our channels is a per-text correlate of workspace
occupancy, it is the red one, and their own construction says why: a
J-lens vector is anchored on a single vocabulary token, and the
workspace contents are, by construction, unspoken \emph{words}, held
active across a span of positions and released before output. That is
approaching the differentiator's profile, token-anchored, deliberately
transient (the ever-recruited basin), recruited in the reported
workspace regime (red-strong basins at 0.21--0.42 of depth in phi-4; the
cross-architecture comparison is normalized and coarse). Taking the
construction seriously also lowers the philosophical temperature: a
workspace whose contents are approached as suspended Lacanian signifiers
carries none of the Cartesian-theater liabilities the consciousness
vocabulary invites (no stage, no viewer, only patterns held in
suspension long enough to bear on the process), and it fits the data
better, since a theater model has to explain why the entire repertoire
of workspace contents is indexed by single tokens of the output
vocabulary.

The pins are then not workspace content but its terminus: the sites
where what was held in suspension is discharged into the text's standing
structure. Three facts converge on this reading: pin basins saturate at
\ensuremath{\approx} 72\% of depth, late in the reported workspace regime;
blue-strong basins recruit later than red-strong basins wherever the
reading is strongest (9/9 phi-4 runs, 7/9 Qwen, 1/9, inverted, in the
base model); and the two channels are mutually exclusive at the top,
top-decile blue and red sets disjoint in 9/9 phi-4 and 8/9 Qwen runs,
the pins sitting in a red hole (Results §3, confound checked there):
where consolidation is maximal, suspension is evacuated. This yields a
concrete cross-instrument prediction: at pin positions the J-lens top-1
persistence should \emph{break}: the pin closes a segment, and the
suspended contents that segment sustained should turn over there at an
elevated rate relative to matched non-pin positions. The prediction is
testable with their released tooling, which has been independently
replicated on an open-weight model; the same bases would enable a second
test, the projection of aggregator core subspaces onto the identified
feature directions, which is beyond the reach of this paper. Their
finding that post-training substantially reshapes J-space contents
meets, finally, a tested result rather than an assumption on our side:
within the one matched lineage, instruction tuning reshapes the reading
(register, fidelity, exclusion) while same-type transfer stands unmoved
(Results §6). If the J-space reshaping is the same training effect, it
belongs to the relational layer tuning rewrites, not to the type-carried
substrate it leaves alone; the rhyme is not yet an identity.

The framing question, finally, can be put mechanistically rather than
philosophically. In the neuroscience it comes from, the workspace
construct earns its empirical content through dissociations, masking and
no-report paradigms separate what cortex demonstrably processes from
what the subject can report. A language model admits no such
dissociation, not for practical reasons but by construction: the
training objective fuses what the system can verbalize with what it
computes with, so verbalizability is not an independent probe of the
computation but its optimization target. The headline convergence of
Gurnee et al.~(the representations the model can report are the ones it
reasons with) is therefore close to analytic where the human finding is
synthetic; and once the dissociation is unavailable, what remains of the
workspace framing is the staging vocabulary itself (ignition, broadcast,
access), none of which does explanatory work that their own operations
do not already do under a thinner description.

The thinner description is available, and it is the one this paper has
been using: sparse directions indexed by vocabulary tokens, held active
over a span of positions, bearing causally on what follows (their
steering and ablation results), discharged into output-tied structure
late in the stack. Stated this way the description never leaves the
mechanistic plane, and it is also, clause for clause, the structural
account of a signifying chain: unspoken signifiers in suspension,
informing the process without being said, bound at punctuating points on
the way out. Between the two framings the surplus commitments all sit on
one side; the chain reading adds no entity, no faculty, and no analogy
to consciousness. Parsimony, for once, favors the psychoanalytic
vocabulary, not because it explains more, but because it imports less.
Note, however: the J-lens basis is constructed by backpropagation from
token likelihood, so its vocabulary tie is partly a fact about the
instrument, and ``the workspace is lexical'' is accordingly
underdetermination plus parsimony, not a demonstration; and the razor
cuts both ways: what it favors is the structural core used in this
paper, suspension, retroactive binding, a material-indifferent binding
function, each operationalized above with stated falsification
conditions, while a reading that claimed more of the theory than the
operations support (a subject, an unconscious, a clinic inside the
model) would be exactly what this paragraph criticizes: framing
outrunning instrument.

A last word on the epistemic status of the frame. The reading proposed
here, the model as a theory of language in the Freudian tradition, is
not as such falsifiable, and does not claim to be: no measurement
refutes a perspective. Nor could falsification operate here in its usual
comparative form, since comparing raw geometry across texts would
destroy the very structure under study (the type-transfer test is the
demonstration). The discipline observed instead is twofold. Each
operationalized clause can fail within a text: pre-registered boundary
directions, the capiton clauses, an annotation each reading either
tracks or does not, and across texts the paper claims only replication
of form, never commensuration of values: what repeats is the shape of
the answer, not a pooled magnitude. Where an anomaly was absorbed rather
than conceded (the two-group reading), the absorption was defended by
the novel predictions it generated, and the first of them has been
cashed: the matched-lineage pair was run, and it corrected the reading's
causal half while confirming its behavioral half (Results §6): tuning
confers the fidelity, the lineage fixes the type-anchoring, and
``de-essentializes'' had to be withdrawn. The persistence break at pins
stands as a stated prediction, not a claim this paper cashes.
Falsifiability, in short, lives at the level of clauses asked of one
text at a time; the perspective answers to the older tribunal of
sterility.

\subsection{Limitations}\label{limitations}

The geometry is correlational throughout: no intervention in this paper
establishes that the model \emph{uses} the structures the channels read,
and all claims are calibrated accordingly. All analyses consume a single
teacher-forced pass; free-running generation may organize differently.
Three models is enough for replication counts, not for factorial claims
about size, architecture, or alignment. The matched-lineage
base/instruct pair, run as a follow-up (Results §6), settles the tuning
attribution within one lineage (register dissociation, clinical fidelity
and channel exclusion reproduce under tuning; type-transfer and
recruitment order do not, and are lineage facts), but one lineage is one
lineage: whether the phi/Qwen transfer regime is itself an effect of
pretraining data, scale or architecture is beyond the reach of a single
matched pair. The type-transfer evidence is likewise within-register:
the corpus's shared types pair texts almost only inside register blocks
(clinical with clinical, technical with technical), which is the case
most favorable to a type essence, and cross-register transfer is
untested. The restriction is principled rather than incidental, since
where structure is text-dependent every added genre multiplies pairs
whose comparison only the trained scaffold mediates, but it caps
generality all the same. The corpus's size belongs in the same register:
thirteen texts is what the design can honestly cover when every ground
truth is fixed by hand, by an expert who read the text, and every claim
sits on a dense, full-resolution scan of every token position. It is a
property of the confirmatory design and not of the instrument, which is
training-free, single-pass and unchanged across checkpoints: the
matched-lineage follow-up of Results §6 was produced by running the same
battery over a new model. Tier-2 clinical annotation is single-analyst,
made on excerpts of longer transcripts the analyst knew in full, and
matched to model output by the authors. A companion caveat conditions
the alignment itself: at the term grain the generic plane is large, so
the Tier-2 anchor is carried by the increment above it, graded clean in
phi-4 and the instruct twin, weaker in the base model, and not
adjudicable in Qwen3-8B, together with the two signed dissociations of
Results §5. Attention, the one model-internal rival, sits in the same
plane (Results §8). The plane does not extend to the region grain, where
a different rival, predictive entropy, matches the channel on one of the
marked disruptions. The base model is simultaneously a limitation and a
datum: Llama-3.1-8B tracks the clinical signifiers least reliably
(blue-operative AUC below chance in 1/12 transcript cells, weakest
throughout), which caps what we can claim about base models and is
itself the low end of the fidelity gradient. Two instrument-level
caveats. First, the aggregator is rank-coupled to the core-subspace size
k (Spearman 0.85--1.00 across runs, \ensuremath{\approx} 0.99 in phi-4),
which invites reading blue as a bare count of active dimensions. It is
not one. The channel is the participation ratio of the final-layer
configuration on those dimensions (Methods), and it runs at a fraction
of the count (pooled median \ensuremath{\approx} 0.37 k, never equal to it,
so a high-k anchor whose variance concentrates in a few directions
scores low). The coupling is monotone, not an identity: it is what lets
k stand in for blue on a rank statistic such as the clinical AUC, but
the magnitude the heatmap renders and the rupture ratios of §5 report is
the occupancy, not the tally. Second, first-position anchors inherit
attention-sink amplification, confounded with texts that open on titles
(Results §8). No headline result depends on either.

\section{Ethics Statement}\label{ethics-statement}

Informed patient consent was granted for the local AI analysis of all
clinical transcripts used in this study. The research protocol and data
handling procedures were reviewed and approved by the responsible ethics
committee.

\section{Acknowledgements}\label{acknowledgements}

We sincerely thank Steffen Brandt and Marie-Louisa Borchardt for the
invitation to Coding.Waterkant, where a major part of this research was
built and freely discussed with the community.

\section{References}\label{references}

\begin{itemize}
\tightlist
\item
  Abdou, Mostafa, Razia S. Sahi, Thomas D. Hull, Erik C. Nook, and
  Nathaniel D. Daw. 2025. ``Leveraging Large Language Models to Estimate
  Clinically Relevant Psychological Constructs in Psychotherapy
  Transcripts.'' \emph{Computational Psychiatry} 9 (1): 187--209.
  \url{https://doi.org/10.5334/cpsy.141}.
\item
  Abnar, Samira, and Willem Zuidema. 2020. ``Quantifying Attention Flow
  in Transformers.'' In \emph{Proceedings of the 58th Annual Meeting of
  the Association for Computational Linguistics}, 4190--4197.
  \url{https://doi.org/10.18653/v1/2020.acl-main.385}.
\item
  Aghazadeh, Ehsan, Mohsen Fayyaz, and Yadollah Yaghoobzadeh. 2022.
  ``Metaphors in Pre-Trained Language Models: Probing and Generalization
  Across Datasets and Languages.'' Preprint, arXiv.
  \url{https://doi.org/10.48550/ARXIV.2203.14139}.
\item
  Aroyo, Lora, and Chris Welty. 2015. ``Truth Is a Lie: Crowd Truth and
  the Seven Myths of Human Annotation.'' \emph{AI Magazine} 36 (1):
  15--24.
\item
  Chen, Guoxuan, Han Shi, Jiawei Li, et al.~2024. ``SepLLM: Accelerate
  Large Language Models by Compressing One Segment into One Separator.''
  Preprint, arXiv. \url{https://doi.org/10.48550/ARXIV.2412.12094}.
\item
  Fuoli, Matteo, Weihang Huang, Jeannette Littlemore, Sarah Turner, and
  Ellen Wilding. 2025. ``Metaphor Identification Using Large Language
  Models: A Comparison of RAG, Prompt Engineering, and Fine-Tuning.''
  Preprint, arXiv. \url{https://doi.org/10.48550/ARXIV.2509.24866}.
\item
  Geal, Robert. 2025. ``A Large Language Model Is Structured Like the
  Unconscious: The (Ordinary) Perverse Psychosis of AI.'' \emph{European
  Journal of Psychoanalysis} 12 (1).
  \url{https://www.journal-psychoanalysis.eu/articles/a-large-language-model-is-structured-like-the-unconscious-the-ordinary-perverse-psychosis-of-ai/}
\item
  Gurnee, Wes, Nicholas Sofroniew, Adam Pearce, et al.~2026.
  ``Verbalizable Representations Form a Global Workspace in Language
  Models.'' \emph{Transformer Circuits Thread}.
  \url{https://transformer-circuits.pub/2026/workspace/}.
\item
  Heimann, Marc, and Anne-Friederike Hübener. 2024. ``The Extimate Core
  of Understanding: Absolute Metaphors, Psychosis and Large Language
  Models.'' \emph{AI \& Society}. Advance online publication.
  \url{https://doi.org/10.1007/s00146-024-01971-7}.
\item
  Heimann, Marc, and Anne-Friederike Hübener. 2025. ``Circling the Void:
  Using Heidegger and Lacan to Think about Large Language Models.''
  \emph{Cognitive Systems Research} 91: 101349.
  \url{https://doi.org/10.1016/j.cogsys.2025.101349}.
\item
  Heimann, Marc. 2026a. ``Freudian AI? Transformer Models as a Proof of
  Concept for a Central Hypothesis in Freudian Theory.'' \emph{Lacunae:
  APPI International Journal for Lacanian Psychoanalysis}, no. 29.
\item
  Heimann, Marc. 2026b. ``Violent Hermeneutics: AI and the Weak
  Technical Event.'' In \emph{EVENT and Its Mediation: Philosophical,
  Religious Studies, Literary and Cultural Theoretical Perspectives},
  edited by M. Nyírő, Z. Lurcza, and P. Makai. University of Miskolc
  Press, Faculty of Humanities and Social Sciences. In print.
\item
  Heimann, Marc. 2026c. ``When the Machine Free Associates:
  Psychoanalysis in the Age of AI.'' In Salon ``Problematizing
  Technology and `AI','' \emph{European Journal of Psychoanalysis}.
\item
  Huh, Minyoung, Brian Cheung, Tongzhou Wang, and Phillip Isola. 2024.
  ``The Platonic Representation Hypothesis.'' \emph{Proceedings of the
  41st International Conference on Machine Learning}. PMLR 235.
  \url{https://proceedings.mlr.press/v235/huh24a.html}.
\item
  Ichien, Nicholas, Dušan Stamenković, and Keith J. Holyoak. 2024.
  ``Large Language Model Displays Emergent Ability to Interpret Novel
  Literary Metaphors.'' \emph{Metaphor and Symbol} 39 (4): 296--309.
\item
  Jahromi, Mohammad N. S., Satya M. Muddamsetty, Asta Sofie Stage
  Jarlner, Anna Murphy Høgenhaug, Thomas Gammeltoft-Hansen, and Thomas
  B. Moeslund. 2024. ``SIDU-TXT: An XAI Algorithm for NLP with a
  Holistic Assessment Approach.'' \emph{Natural Language Processing
  Journal} 7 (June): 100078. \url{https://doi.org/10.1016/j.nlp.2024.100078}.
\item
  Jia, Mumin, and Jairo Diaz-Rodriguez. 2026. ``Unsupervised Text
  Segmentation via Kernel Change-Point Detection on Sentence
  Embeddings.'' Preprint, arXiv.
  \url{https://doi.org/10.48550/ARXIV.2601.18788}.
\item
  Kazmierczak, Rémi, Steve Azzolin, Eloïse Berthier, et al.~2024.
  ``Benchmarking XAI Explanations with Human-Aligned Evaluations.''
  Preprint, arXiv. \url{https://doi.org/10.48550/ARXIV.2411.02470}.
\item
  Kramer, Oliver. 2025. ``Conceptual Metaphor Theory as a Prompting
  Paradigm for Large Language Models.'' Preprint, arXiv.
  \url{https://doi.org/10.48550/ARXIV.2502.01901}.
\item
  Lacan, Jacques. 1993. \emph{The Psychoses}. Vol. 3, \emph{The Seminar
  of Jacques Lacan}. W.W. Norton \& Company.
\item
  Lacan, Jacques. 2006. \emph{Écrits: The First Complete Edition in
  English}. Translated by Bruce Fink. W.W. Norton \& Company.
\item
  Magee, Liam, Vanicka Arora, and Luke Munn. 2023. ``Structured like a
  Language Model: Analysing AI as an Automated Subject.'' \emph{Big Data
  \& Society} 10 (2). \url{https://doi.org/10.1177/20539517231210273}.
\item
  Peng, Runyu, Ruixiao Li, Mingshu Chen, Yunhua Zhou, Qipeng Guo, and
  Xipeng Qiu. 2026. ``How Attention Sinks Emerge in Large Language
  Models: An Interpretability Perspective.'' Preprint, arXiv.
  \url{https://doi.org/10.48550/ARXIV.2603.06591}.
\item
  Plank, Barbara. 2022. ``The `Problem' of Human Label Variation: On
  Ground Truth in Data, Modeling and Evaluation.'' \emph{Proceedings of
  the 2022 Conference on Empirical Methods in Natural Language
  Processing}, 10671--82. \url{https://aclanthology.org/2022.emnlp-main.731/}.
\item
  Rudman, William, Catherine Chen, and Carsten Eickhoff. 2023. ``Outlier
  Dimensions Encode Task Specific Knowledge.'' \emph{Proceedings of the
  2023 Conference on Empirical Methods in Natural Language Processing},
  14596--605. \url{https://doi.org/10.18653/v1/2023.emnlp-main.901}.
\item
  Rudman, William, and Carsten Eickhoff. 2023. ``Stable Anisotropic
  Regularization.'' Preprint, arXiv.
  \url{https://doi.org/10.48550/ARXIV.2305.19358}.
\item
  Solbiati, Alessandro, Kevin Heffernan, Georgios Damaskinos, Shivani
  Poddar, Shubham Modi, and Jacques Cali. 2021. ``Unsupervised Topic
  Segmentation of Meetings with BERT Embeddings.'' Preprint, arXiv.
  \url{https://doi.org/10.48550/ARXIV.2106.12978}.
\item
  Timkey, William, and Marten van Schijndel. 2021. ``All Bark and No
  Bite: Rogue Dimensions in Transformer Language Models Obscure
  Representational Quality.'' Preprint, arXiv.
  \url{https://doi.org/10.48550/ARXIV.2109.04404}.
\item
  Uma, Alexandra N., Tommaso Fornaciari, Dirk Hovy, Silviu Paun, Barbara
  Plank, and Massimo Poesio. 2021. ``Learning from Disagreement: A
  Survey.'' \emph{Journal of Artificial Intelligence Research} 72:
  1385--1470.
\item
  Viswanathan, Karthik, Yuri Gardinazzi, Giada Panerai, Alberto
  Cazzaniga, and Matteo Biagetti. 2025. ``The Geometry of Tokens in
  Internal Representations of Large Language Models.'' Preprint, arXiv.
  \url{https://doi.org/10.48550/ARXIV.2501.10573}.
\item
  Yan, Yu, Sheng Sun, Zenghao Duan, et al.~2025. ``From Benign Import
  Toxic: Jailbreaking the Language Model via Adversarial Metaphors.''
  Preprint, arXiv. \url{https://doi.org/10.48550/ARXIV.2503.00038}.
\end{itemize}

\end{document}